\documentclass[lettersize,journal]{IEEEtran}
\usepackage{amsmath,amsfonts}
\usepackage{algorithmic}
\usepackage{array}
\usepackage[caption=false,font=normalsize,labelfont=sf,textfont=sf]{subfig}
\usepackage{textcomp}
\usepackage{stfloats}
\usepackage{url}
\usepackage{verbatim}
\usepackage{graphicx}
\usepackage{multirow}
\usepackage{xcolor}

\def\BibTeX{{\rm B\kern-.05em{\sc i\kern-.025em b}\kern-.08em
    T\kern-.1667em\lower.7ex\hbox{E}\kern-.125emX}}
\usepackage{balance}
\begin{document}
\title{RoNet: Rotation-oriented Continuous Image Translation}
\author{Yi Li*, Xin Xie, Lina Lei, Haiyan Fu, Yanqing Guo
\thanks{Corresponding author: 

\ \ \ \ \ \ \ \ \ \ \ \ \ \ \ \ \ \ \ \ \ liyi@dlut.edu.cn 

Other authors: 

\ \ \ \ \ \ \ \ \ \ \ \ \ \ \ \ \ \ \ \ \ shelsin@mail.dlut.edu.cn

\ \ \ \ \ \ \ \ \ \ \ \ \ \ \ \ \ \ \ \ \ leilina@mail.dlut.edu.cn

\ \ \ \ \ \ \ \ \ \ \ \ \ \ \ \ \ \ \ \ \ fuhy@dlut.edu.cn,

\ \ \ \ \ \ \ \ \ \ \ \ \ \ \ \ \ \ \ \ \ guoyq@dlut.edu.cn
}}


\maketitle

\begin{abstract}
The generation of smooth and continuous images between domains has recently drawn much attention in image-to-image (I2I) translation. Linear relationship acts as the basic assumption in most existing approaches, while applied to different aspects including features, models or labels. However, the linear assumption is hard to conform with the element dimension increases and suffers from the limit that having to obtain both ends of the line. In this paper, we propose a novel rotation-oriented solution and model the continuous generation with an in-plane rotation over the style representation of an image, achieving a network named \textit{RoNet}. A rotation module is implanted in the generation network to automatically learn the proper plane while disentangling the content and the style of an image. To encourage realistic texture, we also design a patch-based semantic style loss that learns the different styles of the similar object in different domains. We conduct experiments on forest scenes (where the complex texture makes the generation very challenging), faces, streetscapes and the iphone2dslr task. The results validate the superiority of our method in terms of visual quality and continuity. 
\end{abstract}


\begin{figure}[t]
  \centering
   \includegraphics[width=1.0\linewidth]{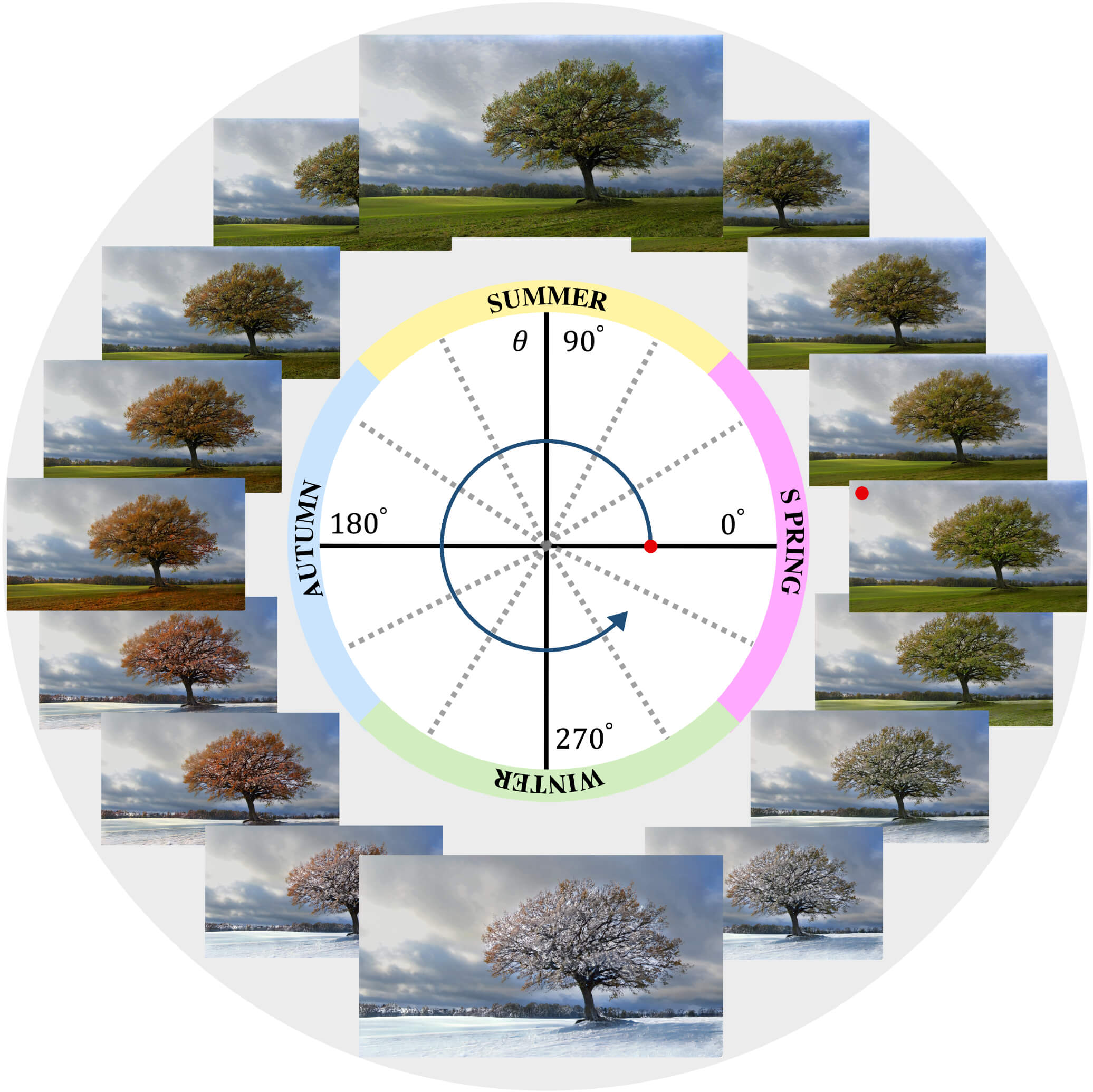}
   \caption{\textbf{The turning wheel of four seasons} generated by RoNet with the single input (on the right labeled with the red dot).}
   \label{fig:wheel}
\end{figure}

\begin{figure*}[t]
  \centering
   \includegraphics[width=1.0\linewidth]{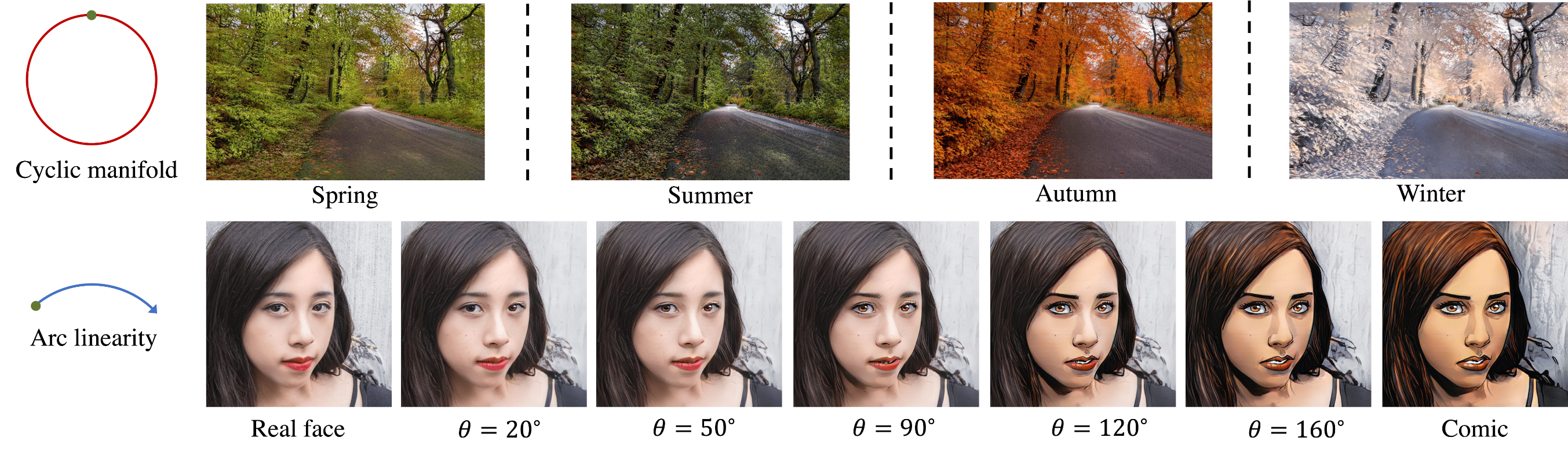}
   \caption{\textbf{The high definition results of RoNet.} Images in one row are generated with a single source image by setting different rotation angles $\theta$. More results are presented in Sec.\ref{Experiments}.}
   \label{fig:show}
\end{figure*}
\section{Introduction}
\IEEEPARstart{I}{mage}-to-image (I2I) translation \cite{isola2017image}, also known as image translation, learns to map an image from the source domain to an image in the target domain.
In the process, an image is roughly decomposed (explicitly or implicitly) into two components: the content (which is domain-invariant) is expected to remain the same after translation, while the style (which is domain-variant) refers to the changes during translation.
Depending on the concrete tasks, I2I translation can benefit a wide range of applications involving portrait animation \cite{pumarola2020ganimation}, photo enhancement \cite{zhu2017unpaired,huang2018multimodal,chen2019homomorphic}, painting style transfer \cite{lee2018diverse,li2018closed,chen2021diverse}, domain adaptation \cite{hoffman2018cycada,murez2018image,li2019bidirectional} and face synthesis \cite{li2020disentangled,han2021disentangled}. 
Despite the impressive progress in the past years, it remains challenging to obtain smooth and continuous translation results.

\begin{figure}[t]
  \centering
   \includegraphics[width=0.85\linewidth]{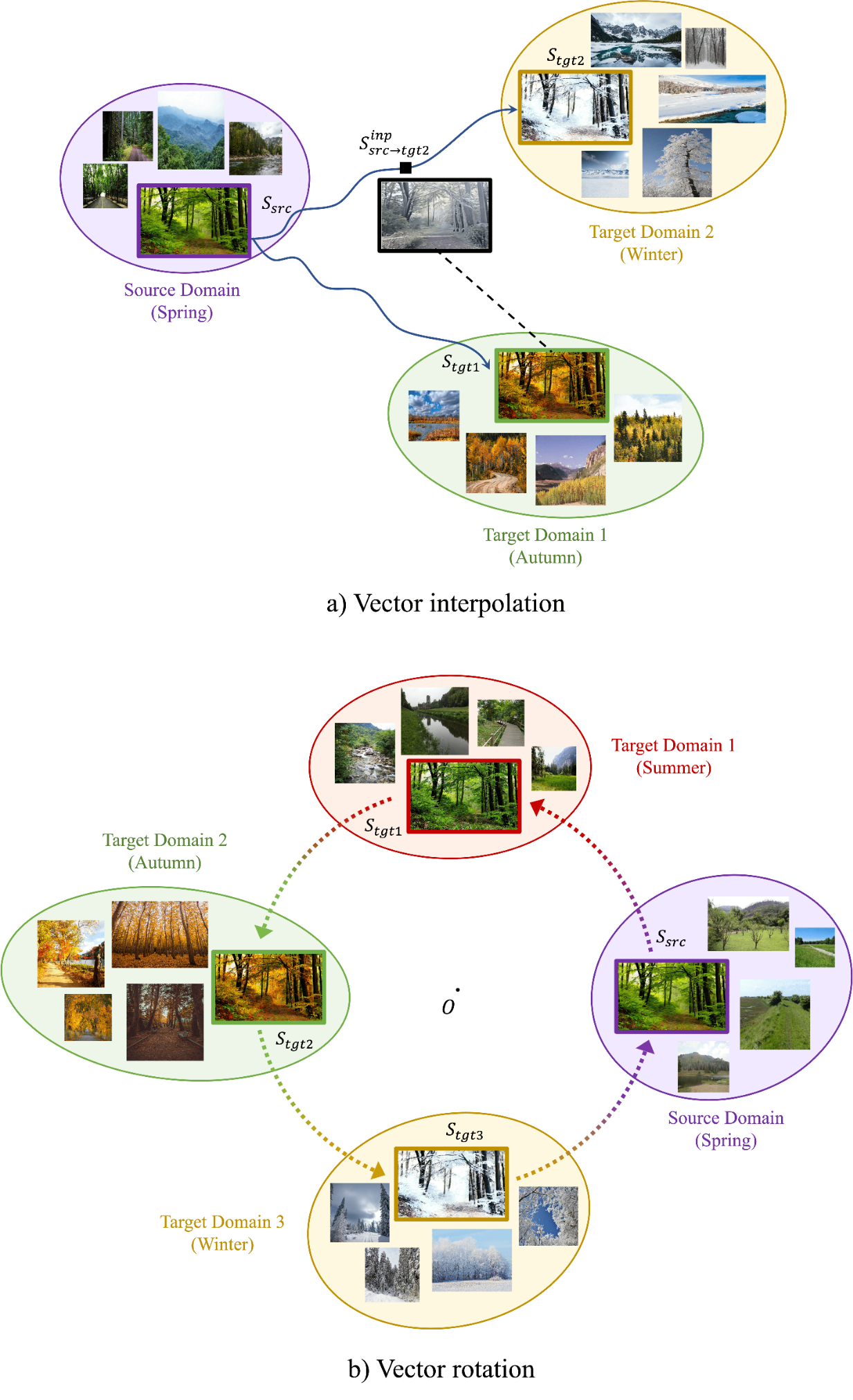}
   \caption{\textbf{Visualized difference between vector rotation and interpolation}.}
   \label{fig:difference}
\end{figure}

Recently, some approaches have explored to utilize linear interpolation to realize continuous translation.
The linear assumption may be imposed on the learned features \cite{xiao2017dna, zhang2019multi}, the trained models \cite{wang2019deep} or the domain labels \cite{gong2019dlow, gong2021analogical}.
The linear manifold assumption is intuitive but has limitations in at least three aspects.
(1) Elements (feature/model/label) of the source domain and the target domain are both necessary to realize the linear interpolation. 
Take the model interpolation in \cite{wang2019deep} as an example, one has to train multiple models to deal with the multi-domain translation.
(2) The elements (feature/model/label) tends to have high dimensions. With the dimension increases, the relationship between two domains becomes harder to conform to the linear assumption, especially when there is a wide gap between two domains.
Even for labels, there are complex scenarios like cyclic translations, e.g., the turning wheel of seasons as presented in Fig.\ref{fig:wheel}, generated by RoNet.
(3) Suppose there are the source representation $S_{src}$ and the target representations $S_{tgt1}$ and $S_{tgt2}$ from two different domains, the fused representation $S^{inp}_{src \rightarrow tgt2}$ is usually defined as $\alpha S_{src} + (1-\alpha) S_{tgt2}$. As illustrated in Fig.\ref{fig:difference} (a), the typical intermediate representation interpolation will sacrifice the expression ability of the fused element in the multi-domain image translation. For example, we cannot obtain the image with autumn style by the representation interpolation from the spring domain to winter domain.

Based on the analysis, this paper proposes to realize continuous translation with an in-plane rotation of the style representation.
Different from the intuitive linear interpolation, we assume the domains to distribute on a circle in a super-plane and learn the super-plane automatically. 
We accordingly propose RoNet that built on a generation network that explicitly disentangles the content and the style of an image.
The method possesses the following advantages.
It is capable of multi-domain generation with a single input image because the domain relationship is embedded in the rotation angle.
Although provided with the cyclic manifold, the rotation plane is learned automatically in an end-to-end manner, making the network be appropriate for not only periodic translation like season shifting but also general translation tasks like $real \; face \to comic \; portrait$ and $iphone \to dslr$.
In the rotation model, the domain style is captured by the vector direction.
When translating from one domain to another, we merely modify the vector direction to the target domain while keeping the magnitude unchanged as presented in Fig.\ref{fig:difference} (b), reserving the expression ability after translation. More specifically, we can generate the continuous images with the style of four seasons by the representation rotation.

We conduct experiments on various translation tasks including season shifting in forests, $real \; face \to comic \; portrait$, solar day shifting of streetscapes and $iphone \to dslr$.
Among them, the translation of the forest scene is quite challenging due to the extremely complex texture of trees.
The results of existing methods deteriorate a lot in this task, shown in Sec.\ref{Experiments}.
To this end, we design a patch-based semantic style loss by focusing effectively on the style nuances in the matched patches across domains.
Compared with other approaches, RoNet produces the most realistic results on various tasks and achieves the superior quantitative performance.

Our main contributions are summarized as follows.
\begin{itemize}
\item To achieve continuous I2I translation, we propose a novel rotation-oriented mechanism which embeds the style representation into a plane and utilizes the rotated representation to guide the generation. 
RoNet is accordingly implemented to learn the rotation plane automatically while disentangling the content and the style of an image simultaneously.
\item To produce realistic visual effects on challenging textures like trees in forests, we design a patch-based semantic style loss.
It first matches the patches from different domains and then learns the style difference with high pertinency.
\item Experiments on various translation scenarios are conducted, including season shifting in forests, $real \; face \to comic \; portrait$, solar day shifting of streetscapes and $iphone \to dslr$. 
With the guidance of the rotation, RoNet successfully generates realistic as well as continuous translation results with a single input image.
\end{itemize}

\section{Related Work} 
As soon as the seminal work of Image-to-Image translation \cite{isola2017image} was proposed, it showed the excellent performance. Building on that basis, several methods \cite{choi2020stargan, choi2018stargan, karras2019style, karras2020analyzing, karras2021alias, SANet, AdaAttn, StyTR, theiss2022unpaired} are designed to further achieve more surprising effect in three main ways as follows. 

\subsection{Disentangled Representations}
Disentanglement is defined as the act of releasing from a snarled or tangled condition, which is a common tool to extract the high-dimensional features in latent space. Actually, many unsupervised methods \cite{lee2018diverse, huang2018multimodal, jiang2020tsit, park2020contrastive, park2020swapping} utilize disentanglement to capture the content and style feature via the encoder, achieving excellent image-to-image transltaion. In addition, multi-domain image translation \cite{choi2020stargan, choi2018stargan} is accomplished by building relative transformations between style representations in different domains or inputing the style encoder extra information about domain label in semi-supervised training ways. With the guidance of style features, image synthesis \cite{karras2019style, karras2020analyzing, karras2021alias} is allowed to be controllable. With the development of neural networks, generated images become more and more vivid enough to fool the discriminator in GANs \cite{2014Generative} or even some industry experts. However, it is hard for most existing methods to make ideal disentanglement, might causing semantic flipping \cite{theiss2022unpaired}. Recent works \cite{liu2019few, saito2020coco} exploit disentanglement for few-shot generalization capabilities. Besides, the technology of disentanglement is frequently used in image editing. There always are latent directional representations which control the changes of one attribute in the image, possessing some latent semantic interpretation. Prior works \cite{InterFaceGAN, L2M_GAN, IndomainGAN} explore the disentanglement of latent dimensions for image editing by training a linear classifier, discovering the meaningful directions for semantic editing and making remarkable results. Voynov and Babenko \cite{voynov2020unsupervised} propose to learn a candidate matrix and a classifier such that the semantic directions in the matrix can be properly recognized by the classifier. Unsupervised methods GANSpace \cite{harkonen2020ganspace} and SeFa \cite{shen2021closed} perform PCA on the sampled data and the weights of generative model respectively to seperate primary semantic directions in the latent space, generating realistic images with target attribute directions. Similarly, ArtIns \cite{xie2022artistic} make advantages of FastICA \cite{hyvarinen1999fast, koldovsky2006efficient} algorithm to obtain each independent style component vector in style feature space, enabling artwork editing. 

\subsection{Style Transfer}
The ultimate goal of style transfer is to generate plausible artworks, preserving the content of the photograph and owning the style of the painting simultaneously. Gatys \emph{et al.} \cite{gatys} were the seminal work to achieve stylization. During the iterative optimization process, the content and style features are fused by calculating the Gram matrix for loss constraints. Similarly, some works \cite{gatys2016preserving, li2017demystifying, risser2017stable} iteratively flexibly combine content and style of arbitrary images, which are time-consuming. For resource-saving and faster stylization, later works \cite{Perceptual_Losses, Texture_Synthesis, StyleBank, ulyanov2017improved, jing2018stroke, wu2022ccpl} turn to the convolutional neural networks (CNNs) and utilize a feed-forward pass to improve the efficiency of stylization. Moreover, migrating multiple styles into one content image \cite{mutlti_style} is completed by conditional instance normalization, generating excellent stylized results and breaking the limitation of learning one specific style. Recently, arbitrary style transfer methods are paid more attention to facilitate efficient applications. WCT \cite{WCT} is proposed to achieve universal style transfer with two transformation steps including whitening and coloring. Huang \emph{et al.} \cite{AdaIN} propose AdaIN, normalizing the mean and variance of each feature map separately, to adaptively combine the content and style. Due to the convenience, a large number of image generation tasks \cite{StyleGAN, choi2020stargan, Drafting} adopt the AdaIN as the first choice to fuse the content and style representations. Based on CNNs, Jing \emph{et al.} \cite{Dynamic} extends the arbitrary style transfer task by introducing dynamic instance normalization. Avatar-Net \cite{Avatar} utilizes a U-net \cite{u_net} to semantically align the content and the style features. Linear \cite{LST} learns a linear transformation according to the content and style features. With the development of attention mechanism \cite{vision_attention, Transformer}, existing methods \cite{SANet, AdaAttn, StyTR} utilize the encoder-transfer-decoder architecture to generate more high-quality artworks. As for photorealistic image stylization, it can be considered as a special kind of image-to-image translation. Recent works \cite{luan2017deep, li2018closed} design the deep-learning approach to faithfully transfer the reference style of natural scenes.

\begin{figure}[t]
  \centering
   \includegraphics[width=0.9\linewidth]{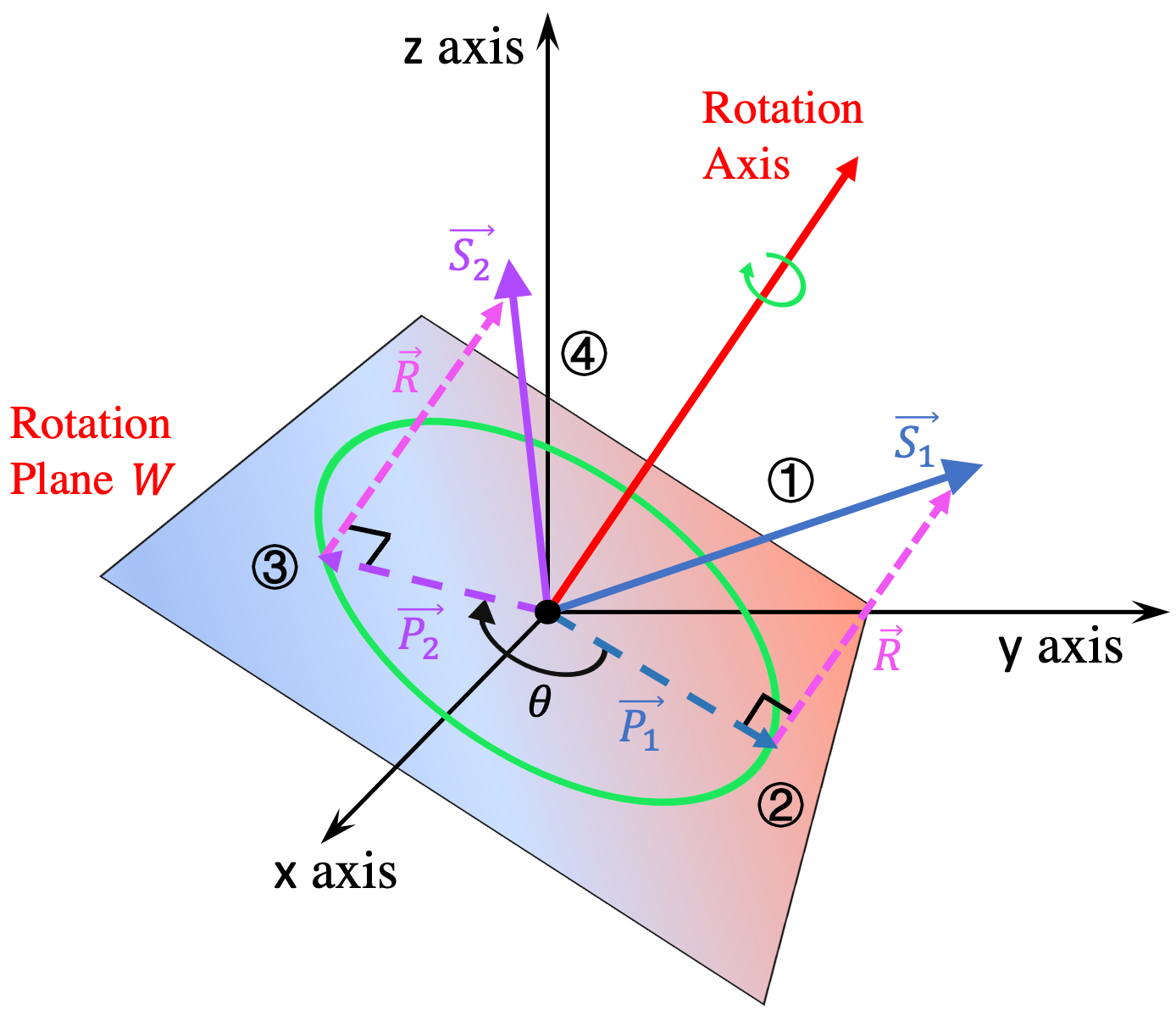}
   \caption{\textbf{Schematic of rotating a style vector from $\vec{S_1} $ to $\vec{S_2}$.} First, $\vec{S_1}$ is mapped onto the rotation plane and obtain $\vec{P_1}+\vec{R} = \vec{S_1}$. Second, rotate $\vec{P_1}$ to $\vec{P_2}$ in the rotation plane. Finally, $\vec{S_2}=\vec{P_2}+\vec{R}$. }
   \label{fig:rotation}
\end{figure}

\subsection{Continuous Image Translation}  
For continuous image variation, feature interpolation \cite{xiao2017dna, zhang2019multi, mao2022continuous} is a common practice to accomplish this task. DRIT \cite{lee2018diverse} and MUNIT \cite{huang2018multimodal} perform continuous interpolation between two style features, while generated images belong to the same domain. StarGAN v2 \cite{choi2020stargan} and SMIT \cite{romero2019smit} mix disentangled style representations, resulting in impressive continuous i2i translation. Besides, continuity can be achieved by model parameter interpolation between two domains \cite{wang2019deep}. Then several methods \cite{gong2019dlow, gong2021analogical, muandet2013domain, li2018deep} generate a continuous sequence of images between two domains by utilizing intermediate domain labels. GANimation \cite{pumarola2018ganimation} adopts a conditional GAN framework \cite{isola2017image}, enabling the continuous generation of examples by inputing the continuous rather than discrete labels at inference time. Relgan \cite{wu2019relgan} introduces the loss interpolation for middle states. In addition, there is rich latent information contained in the underlying dimensions. Chen \emph{et al.} \cite{chen2019homomorphic} have proposed a framework for unpaired i2i translation, generating natural and gradually changing intermediate results by latent space interpolation. CoMoGAN \cite{pizzati2021comogan} relies on naive physics-inspired models to guide the training, learning continuous translations in latent space. However, it is complicated to obtain related physics function for model guidance \cite{pizzati2021comogan} in different domains. Still, linear interpolation \cite{bengio2013better, berthelot2018understanding} is not always valid (e.g. spring to winter include summer and autumn, day to night include dusk).

\begin{figure*}[t]
  \centering
   \includegraphics[width=1.0\linewidth]{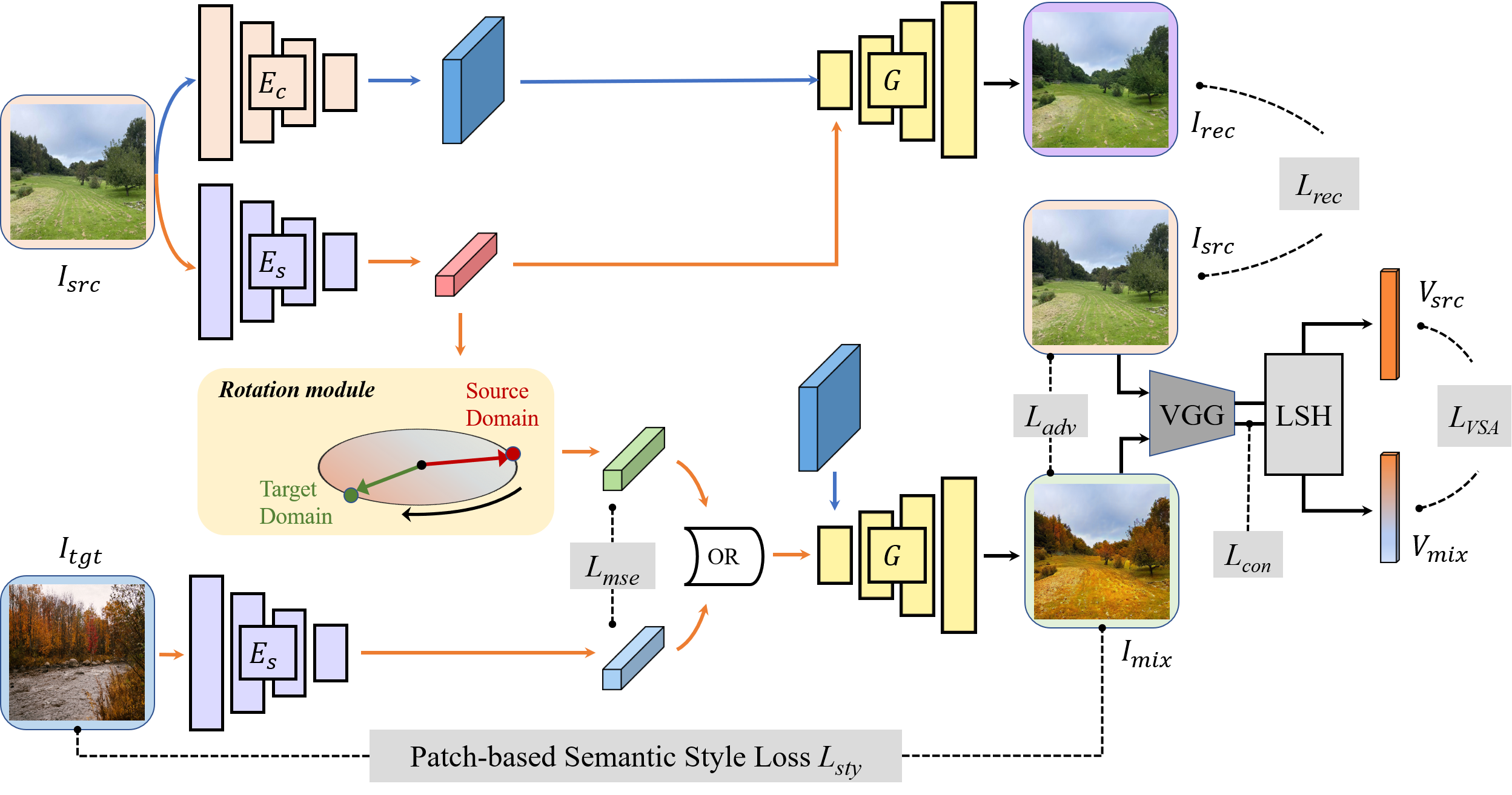}
   \caption{\textbf{Overview of RoNet.} The source image $I_{src}$ is disentangled into the content representation and the style representation by $E_c$ and $E_s$. Under the alternant training of style vector, the style representation of $I_{src}$ is rotated from the source domain to the target domain with the guide of $I_{tgt}$. To further encourage realistic texture, we design a patch-based semantic style loss. }
   \label{fig:framework}
\end{figure*}

\section{Method}
Let $\{\mathcal{I}^i\}_{i=1}^N \in \mathcal{I}$ be the image sets of $N$ different domains, and $\{y_i\}_{i=1}^N \in \mathcal{Y}$ be their corresponding domain labels.
Given an image $\mathbf{I} \in \mathcal{I}$, the goal of RoNet is to generate continuous results across domains that accord with the cyclic manifold, under the guidance of style vector rotation.
Since the style indicates the changes during translation, we employ the style representation to play the rotation role.
Note that the plane is not manually appointed but learned along with the whole network, and the details are illustrated in the following.

\subsection{How to Rotate?}
It is challenging to imagine the rotation of high-dimensional vectors for its spatial complexity. From Euler's rotation theorem we know that any rotation can be expressed as a single rotation with respect to some axes \cite{teoh2005formula}.  
In order to further discuss the rotation process, we define the $n$-dimensional source vector as $S_{1} \in R^{n}$ and let $\theta$ be the angle  to rotate. 
We first project $S_{1}$ onto the $2$-dimensional rotation plane $W$ which is the span of two orthogonal unit vectors $m,n \in R^{n}$:

\begin{equation}
    P_{1} = (S_{1} \cdot m) m + (S_{1} \cdot n) n
    \label{eq1}
\end{equation}
where $P_{1}$ is the projection of $S_{1}$ in $W$. Then we can obtain the rest of $S_{1}$:

\begin{equation}
    R = S_{1} - P_{1}
    \label{eq2}
\end{equation}

where $R$ is the component of $S_{1}$ that is orthogonal to plane $W$. Hence, it is unchanged by the rotation in $W$. Next step, we shall achieve the high-dimensional vector rotation by rotating $P_{1}$ in $W$ as Equation \ref{eq3}, and then mapping the result $P_{2}$ back to plane in which $S_{1}$ lies, which is described as Equation \ref{eq4}.

\begin{equation}
\begin{aligned}
    P_{2} &=Rot_{W,\theta}(P_{1}) \\
    & =[(S_{1} \cdot m)cos\theta-(S_{1} \cdot n)sin\theta]m \\
    &+[(S_{1} \cdot n)cos\theta+(S_{1} \cdot m)sin\theta]n
    \label{eq3}
\end{aligned}
\end{equation}

\begin{equation}
\begin{aligned}
    S_{2} = & P_{2}+R \\
    =S_{1}+\begin{bmatrix} m 
  &n
\end{bmatrix}&\begin{bmatrix}(cos \theta -1)  
  & -sin \theta \\ sin \theta 
  & (cos \theta -1)
\end{bmatrix}\begin{bmatrix}S_{1} \cdot m
 \\ S_{1} \cdot n
\end{bmatrix}
    \label{eq4}
\end{aligned}
\end{equation}

The whole rotation process is shown in the Figure \ref{fig:rotation}. In other words, the vector $S_{2}$ under rotation by $\theta$ is equal to the projection $P_{1}$ of vector $S_{1}$ under rotation by $\theta$ in $W$ plus $(S_{1}-P{1})$. In this work, our target is to discover the rotation plane, allowing style vectors after rotation to be cyclic. So we set two learnable vectors $(\mu,\nu) \in R^{n}$ to model the rotation plane. For specifying the orthogonal unit assumptions, we adopt the Schmidt orthogonalization to obtain the rotation plane:

\begin{equation}
    W=(m,n)=GramSchmidt(\mu,\nu) .
\label{eq5}
\end{equation}

\subsection{RoNet}
The overview of the proposed RoNet is presented in Figure \ref{fig:framework}. 
There are four essential subnets in RoNet which are the content encoder $E_c$, the style encoder $E_s$, the rotation module and the Generator $G$.
Considering it is time- and labour-consuming to obtain the continuous training data, we employ the images in distinct key domains for learning, making the method weakly supervised.
In each round of training, we feed the network with a pair of images $(I_{src}, I_{tgt})$ that are from the source domain and the target domain respectively.
For instance, the source image $I_{src}$ in Figure \ref{fig:framework} is from the \textit{Summer} domain, while the target image $I_{tgt}$ is from the \textit{Autumn} domain.

Given an image, we use the content encoder $E_c$ to extract the domain-invariant representation $C$, and use the style encoder $E_s$ for domain-variant representation $S$.
In a typical exemplar-based I2I translation approach, $I_{src}$ provides the content for the results which should be kept the same, and $I_{tgt}$ specifies the target style of the generated image, i.e., $I_{mix} = G(C_{src}, S_{tgt}) = G(E_c(I_{src}), E_s(I_{tgt}))$.
However, as has been introduced above, RoNet is in charge of both information disentanglement and learning the proper plane for continuously rotation.
Hence we implant a rotation module in the network to find the plane in Equation \ref{eq5}.
Concretely, we first rotate the style vector of the source image $S_{src}$ to the target domain, and then use the rotated vector $S_{rot}$ to generate the image in the target domain $I_{mix} = G(C_{src}, S_{rot})$. 
By alternate training and imposing constraints between $S_{rot}$ and $S_{tgt}$, we learn the rotation plane along with the disentanglement in an end-to-end manner.

In order to extract the content representation and prevent semantic flipping, we adopt normalized content features and segmentation pseudo-labels \cite{theiss2022unpaired} of $I_{src}$ and $I_{mix}$ for better visual quality during generation. 
Different from existing point-to-point translation that usually uses face or animal images in experiments, we apply RoNet in multi-domain scene generation. 
Generation targets as faces hold clear structures, but a scene image tends to consist of stuff without fixed shapes, e.g., sky and grass.
Nevertheless, the content representation is incapable of serving the sketch information for the stuff after rounds of downsampling. 
Thus we complement the generator with content features which can be easily obtained by VGG encoder \cite{simonyan2014very}. On the other hand, the source and target domains always have a large semantic mismatch, suffering from source content corruption. Therefore hypervector is obtained by Vector Symbolic Architectures (VSA) \cite{theiss2022unpaired} to constrain the semantic information of principal components between $I_{src}$ and $I_{mix}$.

\subsection{Loss Optimization}
This section introduces how to build loss functions to optimize the model.

\subsubsection{Adversarial Objective} During the training, the generator takes the content features $C$ and style features $S$ as inputs, learning to generate realistic images via an adversarial loss:
\begin{equation}
\begin{aligned}
    \mathcal{L}_{adv} & = \mathbb{E} \left[\mathrm{log} D_{y_{src}}(I_{src}) \right] \\
    & +\mathbb{E} \left[\mathrm{log} (1-D_{y_{tgt}}(G(C_{src},S_{mix}))) \right]
\label{LossAdv}
\end{aligned}
\end{equation}
where $D_{y}(\cdot)$ means the output of discriminator $D$ corresponding to the domain $y$. Notably, style features $S_{mix}$ are $S_{rot}$ or $S_{tgt}$. Concretely, we input $S_{rot}$ and $S_{tgt}$ alternately to generator for better harmonization of style encoder $E_{s}$ and rotation plane $W$ during training. In certain aspects, such training strategy helps us discover better cyclic manifold and save computation memory.

\subsubsection{Content Preservation} 
For better content representation, we use a pre-trained VGG network \cite{simonyan2014very} to extract content feature maps for computing the content perceptual loss as follows:
\begin{equation}
\begin{aligned}
    \mathcal{L}_{con} = \mathbb{E} \left[\sum_{i}^{cl} \left \| \varphi_{i}(I_{mix}) - \varphi_{i}(I_{src}) \right \|_{2} \right]
\label{LossCon}
\end{aligned}
\end{equation}
where $\varphi_{i}(\cdot)$ denotes the features extracted from the $i$-th layer in a pre-trained VGG network \cite{simonyan2014very} and $\left \| \cdot \right \|_{2}$ denotes Mean Square Error (MSE). Besides, $cl$ represents layers \{$conv4\_1$, $conv5\_1$\}.

\begin{figure}[t]
  \centering
   \includegraphics[width=1.0\linewidth]{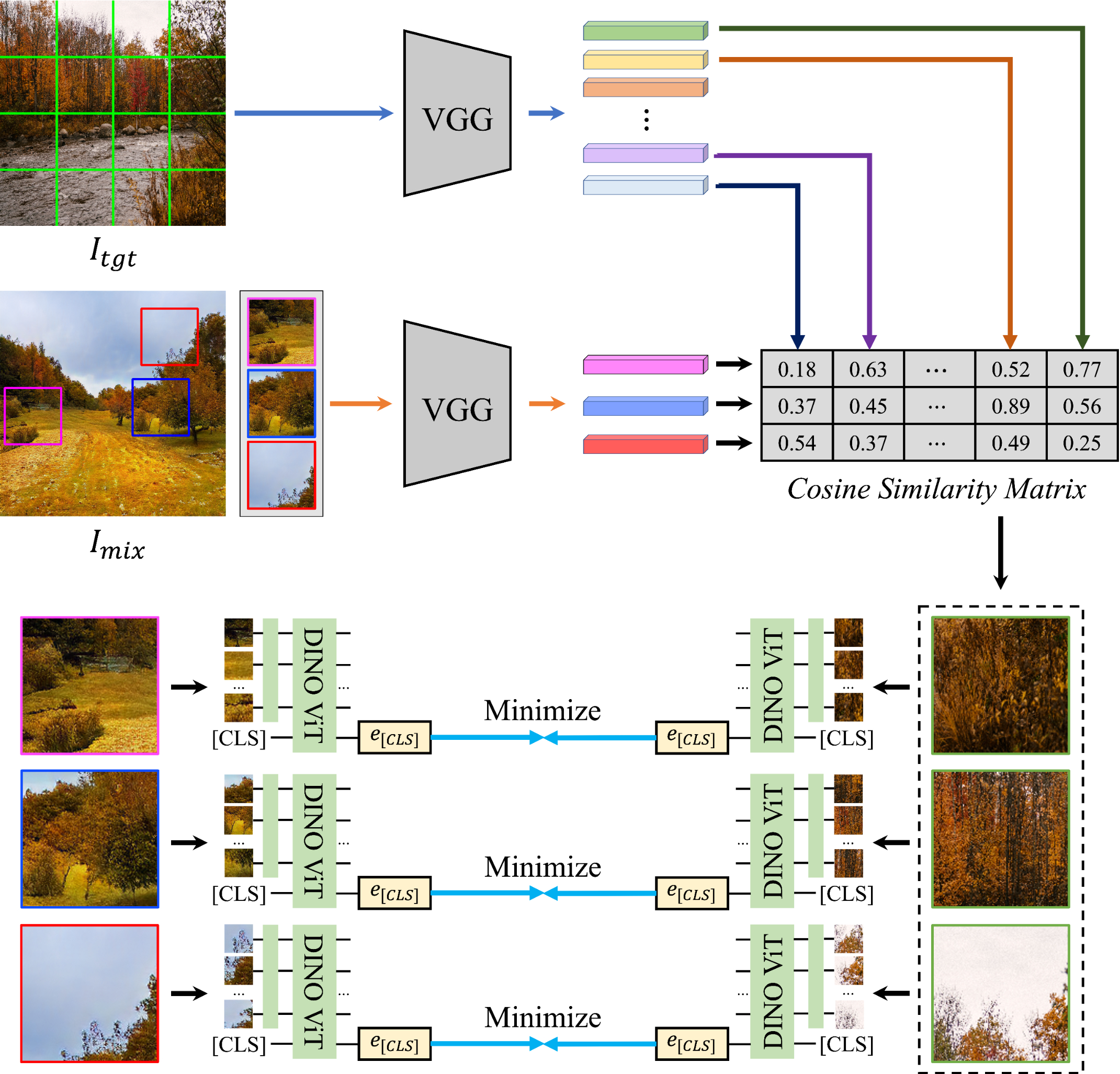}
   \caption{\textbf{Patch-based semantic style loss.} For more realistic texture learning, cosine similarity matrix is calculated for better patch matching according to the style features, encoded from uniform sampling patches of target image $I_{tgt}$ and random sampling patches of generated image $I_{mix}$.}
   \label{fig:PatchLoss}
\end{figure}

\begin{figure*}[t]
  \centering
   \includegraphics[width=1.0\linewidth]{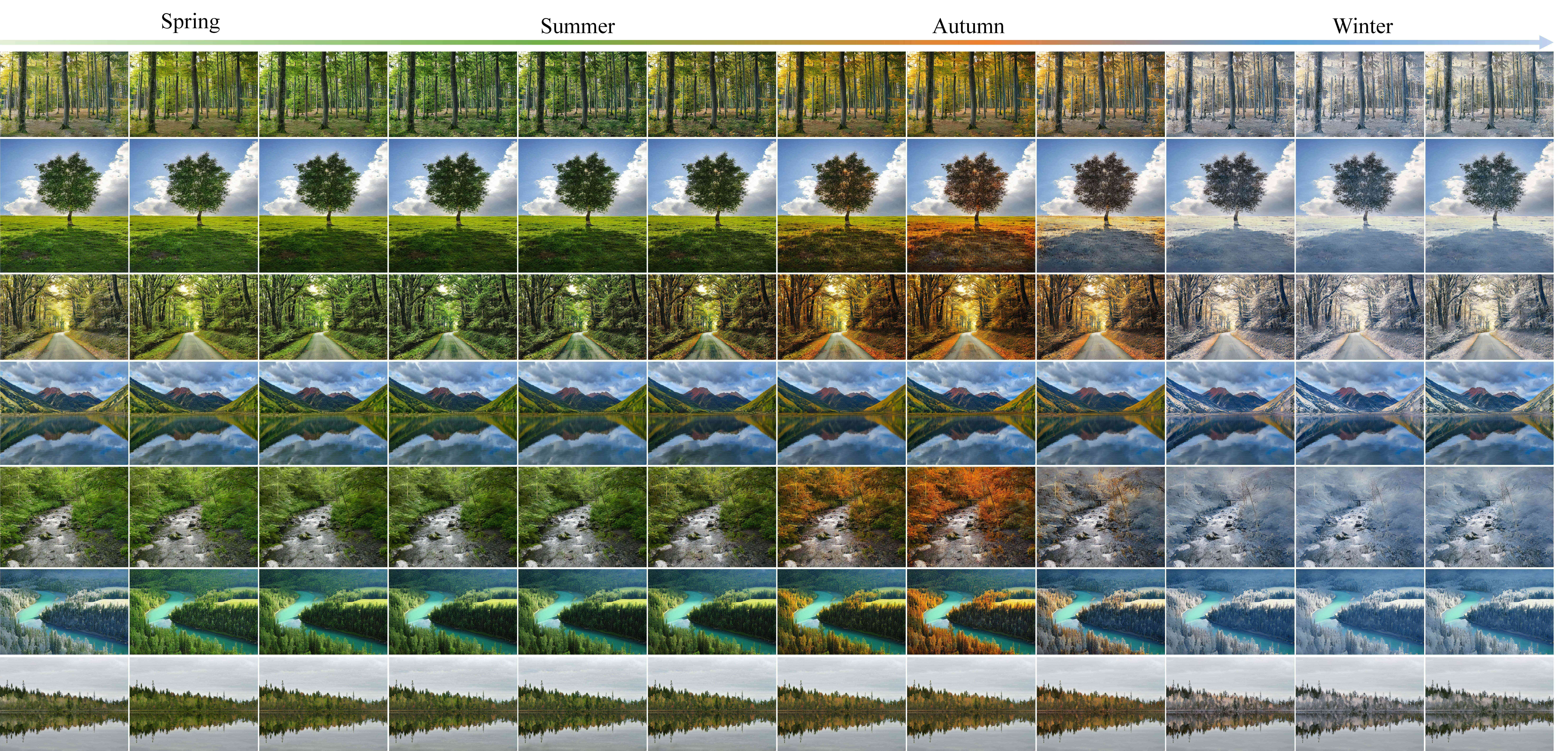}
   \caption{\textbf{Continuous translation from \textit{Spring} to \textit{Winter} generated by RoNet.}}
   \label{fig:season_hq}
\end{figure*}

\subsubsection{VSA-Based Semantic Consistency} Although adversarial and content losses provide generator with powerful constraints, unaligned semantic information between domains results in semantic flipping. Therefore we adapt a VSA-based loss \cite{theiss2022unpaired} to preserve the principal objects of source domain.
\begin{equation}
\begin{aligned}
    \mathcal{L}_{VSA} &= \mathbb{E} \left[1-dist(V_{src}, V_{mix}) \right] 
\label{LossVsa}
\end{aligned}
\end{equation}
where $dist(\cdot, \cdot)$ is the cosine distance. $V_{src}$ and $V_{mix}$ are hypervectors to present the semantic features of $I_{src}$ and $I_{mix}$ respectively, obtained by projecting image features in locality sensitive hashing (LSH) \cite{neubert2019introduction, theiss2022unpaired}.
 
\subsubsection{Patch-based Semantic Matching} Facilitating style encoder $E_{s}$ to capture the realistic texture, we propose a patch-based semantic style loss. In detail, the target image $I_{tgt}$ is uniformly divided into 16 patches while randomly sampling $N$ patches $B^{N}_{mix}$ from the generated image $I_{mix}$. Then cosine similarity matrix is obtained by calculating the cosine distance of patch-based latent representations from two domains, which are extracted by a pre-trained VGG encoder \cite{simonyan2014very}. According to semantic similarity, we can match $N$ patches $B^{N}_{tgt}$ from $I_{tgt}$, processing analogical semantic as $B^{N}_{mix}$. Thus the same objects from different domains are matched for more accurate style learning. For example, the style of trees in $I_{mix}$ is learned from that in $I_{tgt}$. In this work, we adapt a DINO-ViT model \cite{caron2021emerging} (a Vision Transformer model that has been pre-trained in a self-supervised manner) to obtain the $[CLS]$ token, which contains the semantic style information of the image. Therefore we can ensure style capture by minimizing the $[CLS]$ token distances as shown in Figure \ref{fig:PatchLoss}.
\begin{equation}
\begin{aligned}
    \mathcal{L}_{sty} = \mathbb{E} \left [\sum_{i}^{N} \left \| e^{L}_{[CLS]}(B^{i}_{mix}) - e^{L}_{[CLS]}(B^{i}_{tgt}) \right \|_{2} \right].
\label{LossSty}
\end{aligned}
\end{equation}
where $e^{L}_{[CLS]}$ denotes the last layer $[CLS]$ token of DINO-ViT \cite{caron2021emerging}.

\subsubsection{Style Alignment} In order to make style features flow into the cyclic manifold as much as possible. We conduct a loss to restrict the $S_{rot}$ and $S_{tgt}$, ensuring the style features can be rotated from the
source domain to the target domain.
\begin{equation}
\begin{aligned}
    \mathcal{L}_{mse} = \mathbb{E} \left[\left \| S_{rot} - S_{tgt} \right \|_{2} \right]
\label{LossMSE}
\end{aligned}
\end{equation}
Notably, this loss term only work when the input of generator is $S_{rot}$, which helps to make encoder and rotation plane adapt each other.

\subsubsection{Image Reconstruction} To further guarantee that the generator $G$ learns to preserve the original domain-invariant characteristics, we build a reconstruction loss:
\begin{equation}
\begin{aligned}
    \mathcal{L}_{rec} &= \mathbb{E} \left[ ||G(C_{src},S_{src})-I_{src}||_{1} \right] .
\label{LossRec}
\end{aligned}
\end{equation}
Therefore, the network is trained by minimizing the loss function defined as:
\begin{equation}
\label{total_loss}
\begin{aligned}
\mathcal{L} & = \lambda_{adv}\mathcal{L}_{adv} + \lambda_{con}\mathcal{L}_{con} + \lambda_{VSA}\mathcal{L}_{VSA} \\
 & + \lambda_{rec}\mathcal{L}_{rec} + \lambda_{mse}\mathcal{L}_{mse} + \lambda_{sty}\mathcal{L}_{sty}
\end{aligned}
\end{equation}
where $\lambda_{adv}$, $\lambda_{con}$, $\lambda_{VSA}$, $\lambda_{rec}$, $\lambda_{mse}$, $\lambda_{sty}$ are the hyper-parameters to balance each item.

\begin{figure*}[t]
  \centering
   \includegraphics[width=1.0\linewidth]{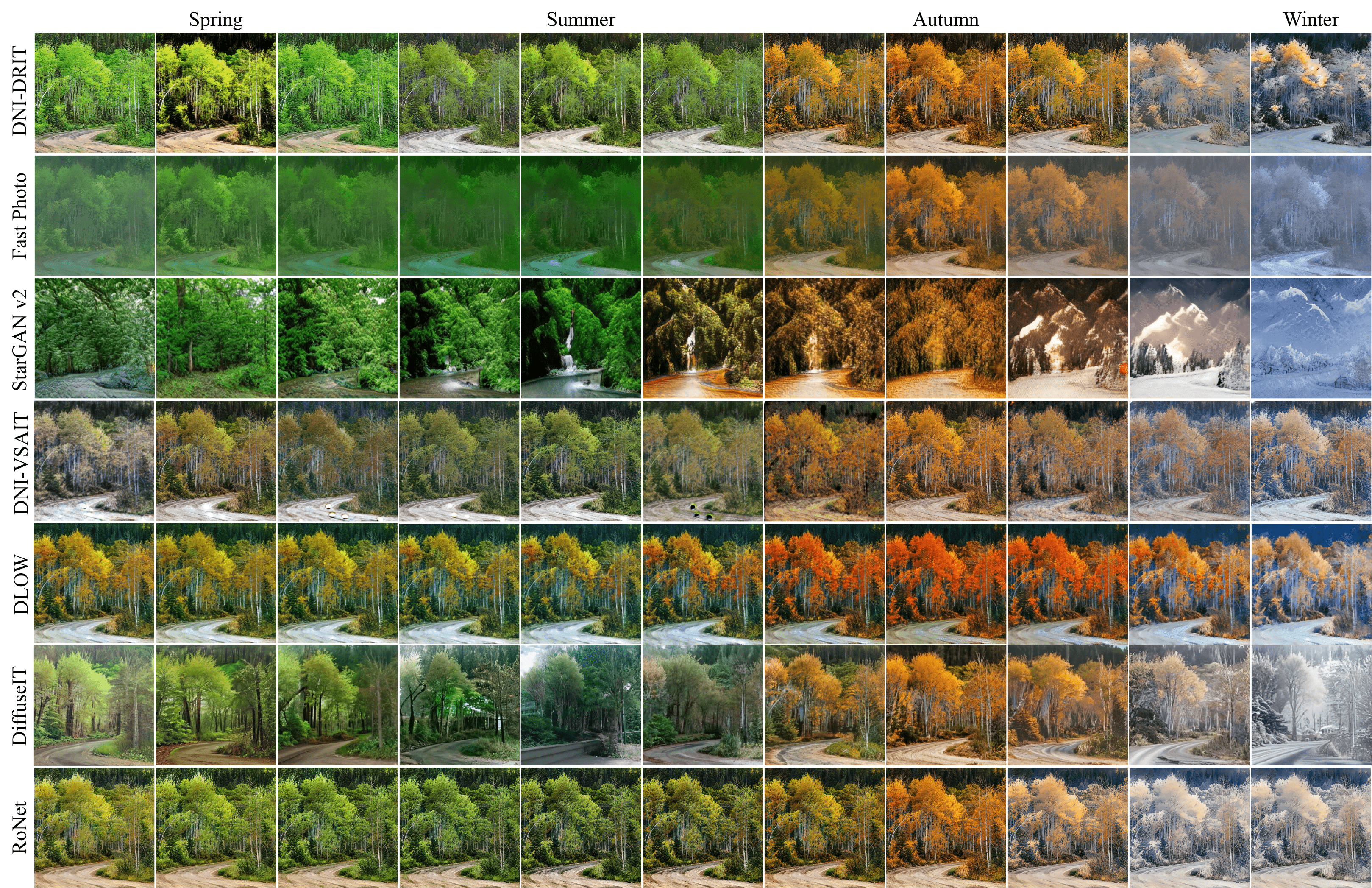}
   \caption{\textbf{Comparison of continuous translation from \textit{Spring} to \textit{Winter}.} The autumn column presents the reconstruction results of the input \textit{Spring} image generated by different approaches. RoNet yields the best results with continuity across domains.}
   \label{fig:season_contrastive}
\end{figure*}

\section{Experiments}
\label{Experiments}
\subsection{Experimental Setting}
\subsubsection{Implementation Details}
During training, we use the ADAM solver \cite{kingma2014adam} with $\beta_{1}=0.0$ and $\beta_{2}=0.99$, and a learning rate of 0.0001 for optimization. Besides, we empirically set the hyperparameters $\lambda_{adv}$, $\lambda_{con}$, $\lambda_{VSA}$, $\lambda_{rec}$, $\lambda_{mse}$ and $\lambda_{sty}$ as 1, 1, 3, 50, 150 and 1 respectively, making the losses balanced. All the experiments are conducted under the environment of Python 3.7.3 and PyTorch 1.7.1 on an Ubuntu 18.04 system with one single 32G Tesla-V00 GPU. In the training stage, all images are randomly cropped into 512 $\times$ 512, while in the inference stage, any image size is supported.

\subsubsection{Datasets}
Our proposed framework is allowed to be applied in different scenes, all experiments are conducted on multiple datasets as follows. 
\begin{enumerate}{}{}
\item \textbf{Season Album}: As for seasonal image translation, we collect 4000 training images and 1000 testing images for each season from \url{flickr.com}. The resolution of all seasonal images is greater than 1024, for keeping the details of seasonal characteristics such as the color of leaves, the texture of snow, etc.
\item \textbf{Comic Face}: We obtain real and comic face dataset from \url{kaggle.com}, which is used to finish real face stylization. There are 1500 real identities and 1500 anime faces for training, and 500 images for testing. 
\item \textbf{Waymo Open Dataset} \cite{sun2020scalability} : There is high-resolution sensor data collected by autonomous vehicles operated by the Waymo Driver in a wide variety of conditions. The scenes of dateset are selected from both suburban and urban area at different moments of the day. The dataset is currently released to making advancements in machine perception and self-driving technology. As for timeshift task, we split clear images into four domains \{day, dusk, night, dawn\} according to Waymo image labels, obtaining train / test sets of 27272 / 7682 images.
\item \textbf{Iphone2dslr Flowers} \cite{zhu2017unpaired} : The dataset is used to map iPhone images with large depth of field to DSLR images with shallow depth of field. There are 1812 / 569 iPhone images and 3325 / 480 DSLR images for training / testing.
\end{enumerate}

\subsubsection{Baselines} We compared our methods with several state-of-the-art methods as follows, which can accomplish the continuous image-to-image translation in different interpolation ways.
\begin{enumerate}{}{}
\item \textbf{StarGAN v2} \cite{choi2020stargan} is a state-of-the-art multi-domain translation network. It can map an input image to multiple defined domains with a single model. We train the model with the public codes released by the authors and use its disentangled style code to enhance the continuous effects with linear interpolation.
\item \textbf{DLOW} \cite{gong2019dlow} realizes continuous translation by generating a continuous sequence of intermediate labels between two domains. In other words, it is a method based on interpolated labels.
\item \textbf{DRIT} \cite{lee2018diverse} 
is able to generate diverse results within a certain domain, but not suitable for multi-domain translation directly. Thus we train the model between every two domains and apply interpolation to the models to obtain the continuous results.
\item \textbf{Fast Photo Style} \cite{li2018closed} is a style transfer method based on disentanglement learning and can generate continuous transfer results via linear interpolation.
\item \textbf{VSAIT} \cite{theiss2022unpaired} is a paradigm for image-to-image translation by setting VSA-based constraints on adversarial learning, achieving continuous image generation by applying the interpolation to the models.
\item \textbf{CoMoGAN} \cite{pizzati2021comogan} achieves cyclic continuous translation with the guidance of physics-inspired models, but is limited in scenarios without decent physical models, such as season transition.
\item \textbf{DiffuseIT} \cite{Gihyun2023DiffuseIT} is a score-based model to accomplish image translation by introducing a loss function to control the diffusion process, but there are instances of content and style leakage in the generated results.
\end{enumerate}

\subsubsection{Evaluation Metrics} We choose the following quantitative evaluation metrics to demonstrate the effectiveness of our proposed framework.
\begin{enumerate}{}{}
\item \textbf{Learned Perceptual Image Patch Similarity} (LPIPS) is based on the VGG \cite{simonyan2014very} and AlexNet \cite{AlexNet} network architectures, evaluating of the distance between image patches. Higher means further different, while lower means more similar.
\item \textbf{Fr\'{e}chet Inception Distance} (FID) \cite{heusel2017gans} is a metric that compares the distribution of generated images with the distribution of a set of real images, by calculating the distance between feature vectors of real and generated images. The FID is the current standard metric for assessing the quality of generative models.
\item \textbf{Kernel Inception Distance} (KID) \cite{borji2019pros} is able to calculate the squared Maximum Mean Discrepancy (MMD) between the Inception representations of the real and generated images, via a polynomial kernel. Similar to the FID \cite{heusel2017gans}, lower values indicate closer distances between the distributions of generated and real data.
\end{enumerate}

\begin{figure*}[t]
  \centering
   \includegraphics[width=1.0\linewidth]{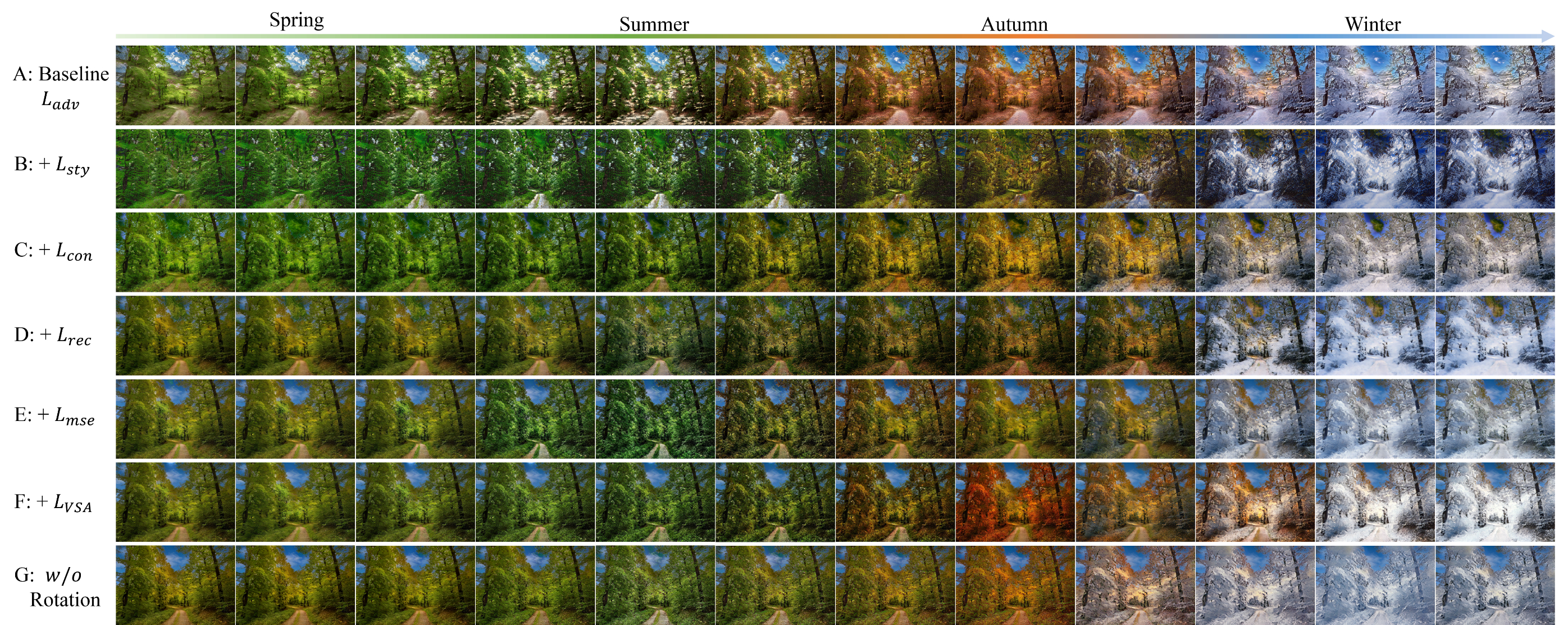}
   \caption{Visual results of ablation studies. The A$\sim$F models are capable of completing the seasonal cycle without the need for reference style images, whereas the G model requires reference style images due to the absence of the rotation module.}
   \label{fig:ablation_show}
\end{figure*}

\begin{figure*}[t]
  \centering
   \includegraphics[width=1.0\linewidth]{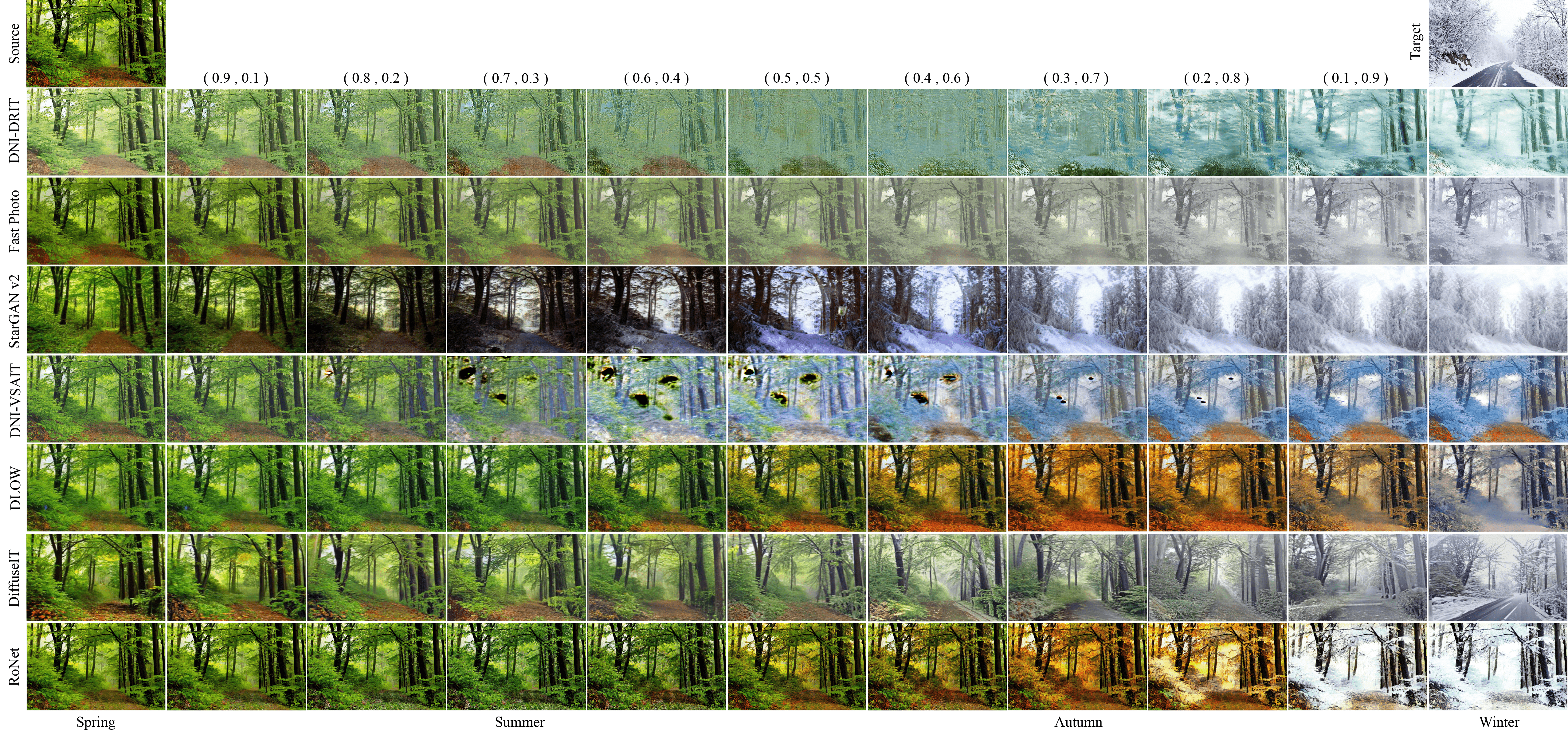}
   \caption{\textbf{Interpolation Effectiveness}. Given the spring image as the source and the winter as the target style, we should obtain the summer and autumn results when doing the interpolation from spring to winter.}
   \label{fig:season_interpolation}
\end{figure*}

\begin{figure*}[t]
  \centering
   \includegraphics[width=1.0\linewidth]{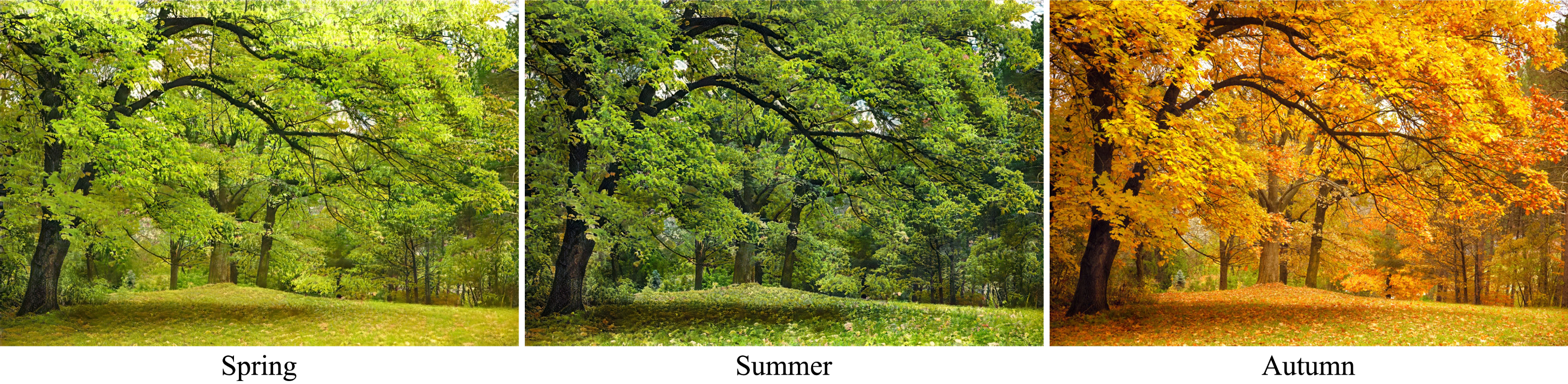}
   \caption{\textbf{high-resolution images} generated by RoNet.}
   \label{fig:high_resolution}
\end{figure*}

\begin{figure*}[t]
  \centering
   \includegraphics[width=1.0\linewidth]{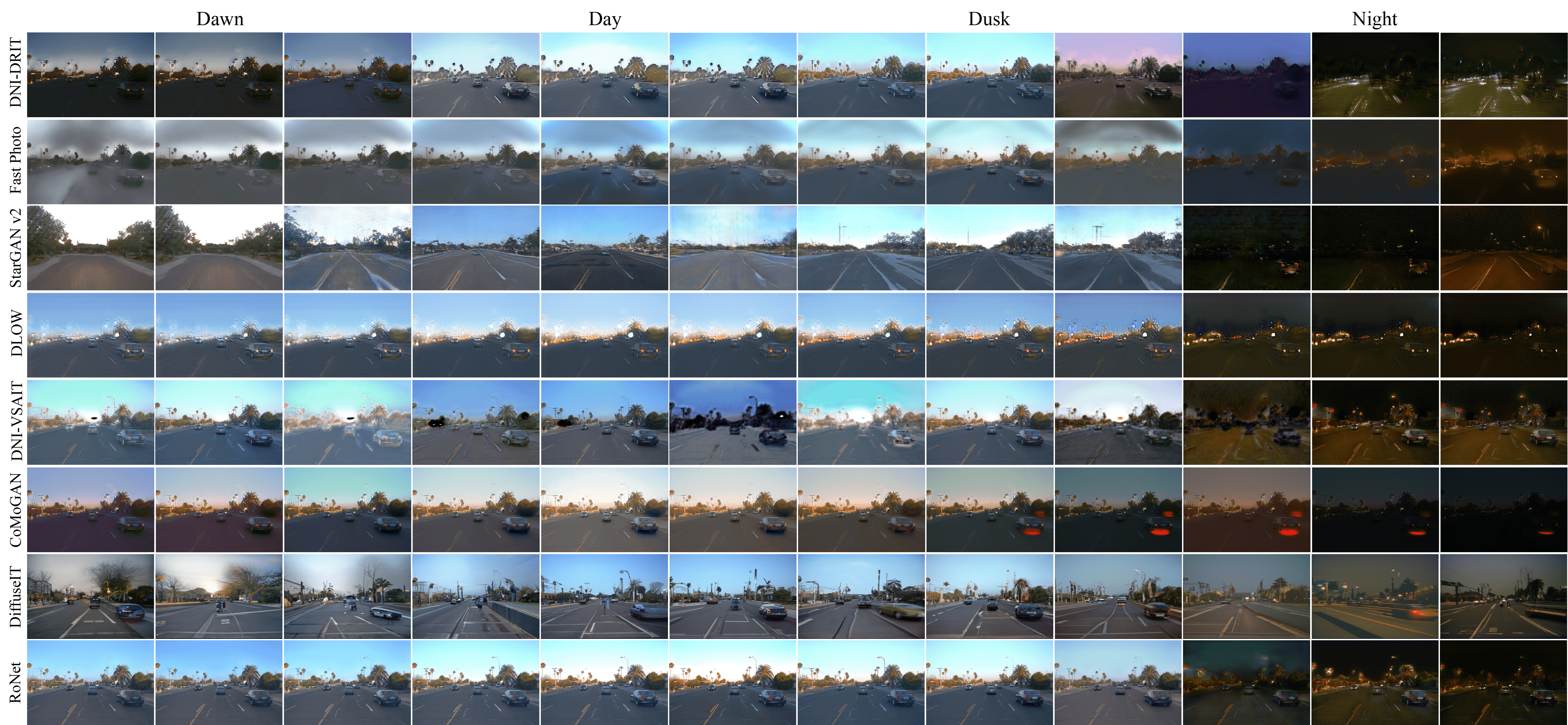}
   \caption{\textbf{Comparison of continuous translation from \textit{Dawn} to \textit{Night}.} The dusk column presents the reconstruction results of the input image generated by different approaches.}
   \label{fig:day_contra}
\end{figure*}

\begin{figure}[t]
  \centering
   \includegraphics[width=1.0\linewidth]{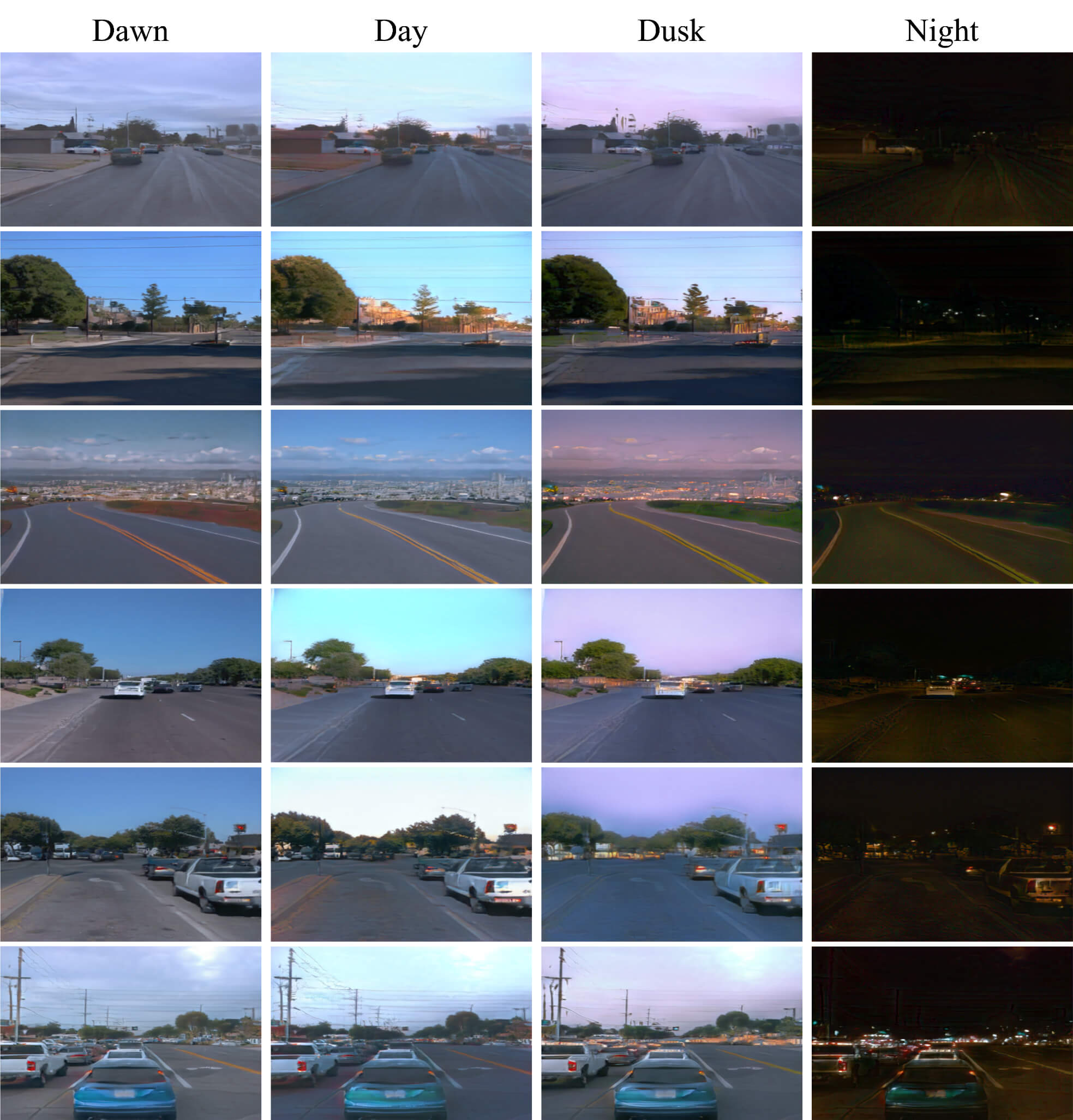}
   \caption{\textbf{Continuous translation from \textit{Dawn} to \textit{Night}.}}
   \label{fig:timeshift}
\end{figure}

\subsection{Season Shifting}
\subsubsection{Continuous Translation} Fig.\ref{fig:season_hq} presents the continuous translation results with high definition.
Please zoom in for more realistic details.
In nature, most plants turn into dark yellow when autumn comes.
RoNet captures the smooth variation successfully with the help of the in-plane rotation.
For typical forest scene as the first row, the complex texture is quite challenging for the net to learn, easily causing artifacts or blur.
It is observed that RoNet preserves the texture well because there is the patch-based semantic style loss.
In the penultimate row, the plants changes with seasons, while the river keeps similar appearance across seasons, which is agree with natural order.
Furthermore, RoNet has the ability of cyclic continuous translation with a single input.
We present raw results in Fig.\ref{fig:wheel} which are also the materials for the corresponding time-lapse demo in the supplementary material.

\subsubsection{Comparison} We compare RoNet with leading approaches in closely related fields, including multi-domain translation, continuous translation based on linear interpolation, and translation based on disentangled representations (StarGAN v2 \cite{choi2020stargan}, DLOW \cite{gong2019dlow}, DRIT \cite{lee2018diverse}, Fast Photo Style \cite{li2018closed}, VSAIT \cite{theiss2022unpaired}, DiffuseIT \cite{Gihyun2023DiffuseIT}). Note that all the approaches are trained and tested under the same protocol.
The comparison is exhibited in Fig.\ref{fig:season_contrastive}.
For StarGAN v2 \cite{choi2020stargan}, we provide the style representations of different target domains to implement the multi-domain translation.
Thus the 2nd, 5th, 8th and last column of StarGAN v2 \cite{choi2020stargan} is guided by certain domain styles, while the rest are the corresponding interpolation results over the style vectors.
For the other four rows, models are trained between \textit{Spring} and \textit{Winter} as they are essentially point-to-point approaches, but applied with different interpolation schemes: DLOW \cite{gong2019dlow} with label interpolation, Fast Photo Style \cite{li2018closed} and DiffuseIT \cite{Gihyun2023DiffuseIT} with representation interpolation, DNI-DRIT \cite{lee2018diverse} and DNI-VSAIT \cite{theiss2022unpaired} with model interpolation. It is observed that RoNet achieves the most appealing results in terms of both visual quality and continuity.

\begin{table*}[]
\renewcommand{\arraystretch}{1.2}
\centering
\caption{Metrics comparison of RoNet and existing approaches.}
\label{tab:contrastive_table}
\begin{tabular}{c|ccccc|ccccc|ccccc}
\hline
\multirow{2}{*}{Method} & \multicolumn{5}{c|}{FID $\downarrow$} & \multicolumn{5}{c|}{LPIPS $\downarrow$} & \multicolumn{5}{c}{KID $\times 10^{-3} \downarrow$} \\ \cline{2-16} 
                        & Spr    & Sum   & Aut   & Win   & mean & Spr    & Sum    & Aut   & Win   & mean  & Spr      & Sum      & Aut      & Win      & mean    \\ \hline
StarGAN v2 \cite{choi2020stargan}             & 45.8   & 74.2  & 81.8  & 66.3  & 67.0 & 0.675  & 0.685  & 0.748 & 0.738 & 0.712 & 34.1     & 39.2     & 43.9     & 35.2     & 38.1     \\
DLOW \cite{gong2019dlow}            & 75.3   & 60.0  & 88.2  & 78.7  & 75.5 & 0.487  & 0.489  & 0.524 & 0.530 & 0.508 & 43.7     & 23.5     & 46.6     & 38.4     & 38.1    \\
DRIT \cite{lee2018diverse}    & 58.3   & 47.2  & 52.8  & 57.9  & 54.1 & 0.306  & 0.306  & 0.363 & 0.457 & 0.357 & 35.8     & 23.4     & 27.1     & 30.2     & 29.1    \\
Fast Photo Style \cite{li2018closed}  & 102.4  & 80.1  & 82.1  & 76.2  & 85.2 & 0.421  & 0.421  & 0.510 & 0.440 & 0.448 & 73.4     & 54.9     & 51.1     & 50.8     & 57.6    \\
VASIT \cite{theiss2022unpaired}   & 59.7      & 52.8     & 60.4     & 69.1     & 60.5    & 0.261      & 0.240      & 0.308     & 0.235     & 0.261     & 37.4        & 30.0        & 34.4        & 41.2        & 35.7       \\
DiffuseIT \cite{Gihyun2023DiffuseIT}   & 60.8      & 49.1     & 55.6     & 49.3     & 53.7    & 0.512      & 0.517      & 0.577     & 0.538     & 0.536     & 41.1        & 24.6        & 30.0        & 22.5        & 29.5       \\
RoNet                   & 55.7      & 43.3     & 57.1     & 55.7     & \textbf{52.9}    & 0.265      & 0.238      & 0.256     & 0.176     & \textbf{0.234}     & 36.8        & 20.5        & 27.8        & 26.2        & \textbf{27.8}       \\ \hline
\end{tabular}
\end{table*}

\begin{table*}[]
\renewcommand{\arraystretch}{1.2}
\centering
\caption{Ablation studies.}
\label{tab:ablation_study}
\begin{tabular}{l|ccccc|ccccc|ccccc}
\hline
\multicolumn{1}{c|}{\multirow{2}{*}{Method}}         & \multicolumn{5}{c|}{FID $\downarrow$} & \multicolumn{5}{c|}{LPIPS $\downarrow$} & \multicolumn{5}{c}{KID $\times 10^{-3} \downarrow$} \\ \cline{2-16} 
                                & Spr    & Sum   & Aut   & Win   & mean & Spr    & Sum    & Aut   & Win   & mean  & Spr      & Sum      & Aut      & Win      & mean    \\ \hline
A: Baseline $\mathcal{L}_{adv}$ & 102.1  & 66.0  & 76.6  & 72.2  & 79.2 & 0.394  & 0.378  & 0.464 & 0.378 & 0.404 & 77.2     & 38.0     & 40.2     & 39.5     & 48.7    \\
B: + $\mathcal{L}_{sty}$        & 65.3   & 66.3  & 72.1  & 74.1  & 69.5 & 0.461  & 0.413  & 0.523 & 0.553 & 0.488 & 47.9     & 52.3     & 39.7     & 39.4     & 44.8    \\
C: + $\mathcal{L}_{con}$        & 58.7   & 53.6  & 59.1  & 63.9  & 58.8 & 0.254  & 0.248  & 0.309 & 0.366 & 0.294 & 34.9     & 29.1     & 28.4     & 37.3     & 32.5    \\
D: + $\mathcal{L}_{rec}$        & 57.1   & 52.9  & 57.9  & 71.0  & 59.7 & 0.248  & 0.218  & 0.233 & 0.350 & 0.262 & 35.7     & 29.2     & 29.7     & 40.5     & 33.8    \\
E: + $\mathcal{L}_{mse}$        & 56.3   & 46.3  & 57.7  & 58.6  & 54.7 & 0.290  & 0.238  & 0.299 & 0.340 & 0.292 & 35.7     & 23.4     & 31.5     & 28.3     & 29.7    \\
F: + $\mathcal{L}_{VSA}$        & 55.7   & 43.3  & 57.1  & 55.7  & \textbf{52.9} & 0.265  & 0.238  & 0.256 & 0.176 & \textbf{0.234} & 36.8     & 20.5     & 27.8     & 26.2     & \textbf{27.8}    \\
G: $w/o$ Rotation     & 59.5   & 54.4  & 65.0  & 67.3  & 61.5 & 0.228  & 0.228  & 0.261 & 0.267 & 0.246 & 41.2     & 31.1     & 39.7     & 35.6     & 36.9    \\ \hline
\end{tabular}
\end{table*}

\begin{table*}[t]
\renewcommand{\arraystretch}{1.2}
\centering
\caption{Metrics comparison of RoNet and existing approaches on the timeshift task.}
\label{tab:time_contrastive}
\setlength{\tabcolsep}{1.9mm}{
\begin{tabular}{c|ccccc|ccccc|ccccc}
\hline
\multirow{2}{*}{Method}                               & \multicolumn{5}{c|}{FID $\downarrow$}                                                                                                      & \multicolumn{5}{c|}{LPIPS $\downarrow$}                                                                                                    & \multicolumn{5}{c}{KID $\times 10^{-3} \downarrow$}                                                                                       \\ \cline{2-16} 
                                                      & Dawn                      & Day                       & Dusk                      & Night                     & mean                       & Dawn                      & Day                       & Dusk                      & Night                     & mean                       & Dawn                      & Day                       & Dusk                      & Night                     & mean                      \\ \hline
StarGAN v2 \cite{choi2020stargan}    & \multicolumn{1}{l}{215.9} & \multicolumn{1}{l}{120.5} & \multicolumn{1}{l}{163.8} & \multicolumn{1}{l}{161.7} & \multicolumn{1}{l|}{165.5} & \multicolumn{1}{l}{0.573} & \multicolumn{1}{l}{0.546} & \multicolumn{1}{l}{0.515} & \multicolumn{1}{l}{0.660} & \multicolumn{1}{l|}{0.573} & \multicolumn{1}{l}{156.3} & \multicolumn{1}{l}{142.0} & \multicolumn{1}{l}{139.9} & \multicolumn{1}{l}{184.2} & \multicolumn{1}{l}{155.6} \\
DLOW \cite{gong2019dlow}             & 184.9                     & 141.2                     & 153.1                     & 134.4                     & 153.4                      & 0.352                     & 0.354                     & 0.340                     & 0.458                     & 0.376                      & 151.3                     & 144.0                     & 143.4                     & 109.8                     & 137.1                     \\
DRIT \cite{lee2018diverse}           & 196.6                     & 156.2                     & 208.2                     & 159.9                     & 180.2                      & 0.407                     & 0.393                     & 0.371                     & 0.627                     & 0.450                      & 153.3                     & 144.1                     & 222.1                     & 118.5                     & 159.5                     \\
Fast Photo Style \cite{li2018closed} & 159.1                     & 162.0                     & 190.0                     & 112.6                     & 155.9                      & 0.317                     & 0.323                     & 0.300                     & 0.523                     & 0.366                      & 114.8                     & 143.2                     & 189.9                     & 94.9                      & 135.7                     \\
VASIT \cite{theiss2022unpaired}      & 114.2                     & 93.2                      & 104.6                      & 100.6                      & 103.4                       & 0.153                     & 0.174                     & 0.174                     & 0.438                     & 0.235                      & 64.0                      & 57.2                      & 77.4                      & 54.7                      & 63.3                     \\
CoMoGAN \cite{pizzati2021comogan}    & 141.2                     & 103.9                     & 141.0                     & 117.9                     & 126.0                      & 0.284                     & 0.283                     & 0.308                     & 0.338                     & 0.303                      & 84.0                      & 67.3                      & 128.1                     & 100.7                     & 95.0                      \\
DiffuseIT \cite{Gihyun2023DiffuseIT}    & 108.3                     & 76.7                     & 103.1                     & 109.5                     & 99.4                      & 0.422                     & 0.424                     & 0.391                     & 0.619                     & 0.464                      & 55.6                      & 55.1                      & 80.6                     & 77.2                     & 67.1                      \\
RoNet                                                 & 107.6                     & 75.1                      & 87.4                     & 93.4                     & \textbf{90.9}                      & 0.178                     & 0.209                     & 0.195                     & 0.212                     & \textbf{0.199}                      & 53.9                      & 31.7                      & 49.6                      & 46.3                      & \textbf{45.4}                      \\ \hline
\end{tabular}}
\end{table*}

\subsubsection{Quantitative Analysis}
In order to evaluate the performance of our model objectively, we report the quantitative metrics of different approaches in Table \ref{tab:contrastive_table}. Three metrics are employed to evaluate these approaches in various views. LPIPS \cite{zhang2018unreasonable} evaluates the structure difference between source images and generated images, FID \cite{heusel2017gans} and KID \cite{borji2019pros} measures the distance between the distribution of the target images and the generated images. For all the three metrics, the smaller the better. Note that all contrastive experiments are conducted with test set in datasets. It is observed that RoNet achieves the lowest performance in all metrics, verifying its superiority over the others.

\subsubsection{Ablation Study} 
To analyse the contribution of each loss function elaborately, we conduct the ablation studies and show the results in Tab.\ref{tab:ablation_study}. The baseline is trained merely by the adversarial loss, then we add each loss function gradually. If we first add the patch-based semantic style loss, the model performance declines much. The reason is that these two loss functions focus more on visual reality but are insufficient for model convergence. With the contribution of the content loss, the reconstruction loss and the MSE loss, the performance increases significantly. Finally, the full model F achieves the best results. In the paper, we report the importance of each loss function as presented in Tab.\ref{tab:ablation_study}. Here, visual effects of each loss function are shown in Fig.\ref{fig:ablation_show}. It is observed that the sky artifacts decrease and semantic style becomes more realistic. In addition, we remove the rotation module to test its impact on the model's performance. The results are shown in Tab.\ref{tab:ablation_study} and Fig.\ref{fig:ablation_show}. The findings demonstrate that the rotation module significantly enhances the model's ability to learn vibrant styles, even without reference style images.

\subsubsection{Interpolation Effectiveness} We apply different interpolation schemes to the state-of-the-art methods: DLOW \cite{gong2019dlow} with label interpolation, StarGAN v2 \cite{choi2020stargan} and Fast Photo Style \cite{li2018closed} with representation interpolation, DNI-DRIT \cite{lee2018diverse} and DNI-VSAIT \cite{theiss2022unpaired} with model interpolation. As shown in Fig.\ref{fig:season_interpolation}, we set the spring image as input source and the winter image as target style. It is observed that StarGAN v2 \cite{choi2020stargan}, DiffuseIT \cite{Gihyun2023DiffuseIT} and Fast Photo Style \cite{li2018closed} suffer from content leak, DNI-DRIT \cite{lee2018diverse} and DNI-VSAIT \cite{theiss2022unpaired} exhibit serious artifacts. However, linear interpolation is not always valid, these methods should generate intermediate results of the summer and autumn seasons. DLOW \cite{gong2019dlow} and RoNet achieve the full seasonal circulation without the target image, but DLOW \cite{gong2019dlow} obtains more style leak.

\subsubsection{High Resolution} Due to the advantageous features of convolutional neural networks, RoNet is capable of producing a plethora of images with varying scales. As illustrated in Fig.\ref{fig:high_resolution}, the images generated by RoNet possess a resolution of $1024 \times 1600$ pixels, which vividly exposes the intricate texture details of leaves.

\subsection{Time Shifting}
\subsubsection{Difference}
Recently, an unsupervised generation network named CoMoGAN \cite{pizzati2021comogan} has been proposed to learn non-linear continuous translations. Despite the unsupervised training manner, CoMoGAN relies on physics-inspired models to guide the learning process. Taking the cyclic translation task of ``day to any time'' as an example, CoMoGAN first renders a daytime image with a color-based model to obtain the images at any time of a day, and then use the series of rendered data as supervision. It makes CoMoGAN stuck in the limitation of seeking the proper physical model. When confronting more challenging tasks such as season transition, CoMoGAN is out of work since it is hard to find a capable physical model for data rendering. However, our proposed framework is not constrained by physical guidance, applying to arbitrary scene domains, such as seasonal variation shown in Fig.\ref{fig:season_hq} and time shifting shown in Fig.\ref{fig:timeshift}. 

\subsubsection{Comparison}
We compare RoNet with several SOTA approaches (StarGAN v2 \cite{choi2020stargan}, DLOW \cite{gong2019dlow}, Fast Photo Style \cite{li2018closed}, DRIT \cite{lee2018diverse}, CoMoGAN \cite{pizzati2021comogan}, VSAIT \cite{theiss2022unpaired}, DiffuseIT \cite{Gihyun2023DiffuseIT}). From Fig.\ref{fig:day_contra} and Tab.\ref{tab:time_contrastive}, it is observed that StarGAN v2 \cite{choi2020stargan} and DiffuseIT \cite{Gihyun2023DiffuseIT} suffer from serious semantic distortion. Fast Photo Style \cite{li2018closed}, DLOW\cite{gong2019dlow}, DNI-VSAIT \cite{theiss2022unpaired} and DNI-DRIT \cite{lee2018diverse} obtain unrealistic results. Since the guidance of physics-inspired function, CoMoGAN \cite{pizzati2021comogan} learns more color- and pixel-wise information. However, our model captures the style from data distribution, such as the street lights in the night and sunsets in the dusk. 

\subsection{Other Tasks}
\begin{figure}[t]
  \centering
   \includegraphics[width=1.0\linewidth]{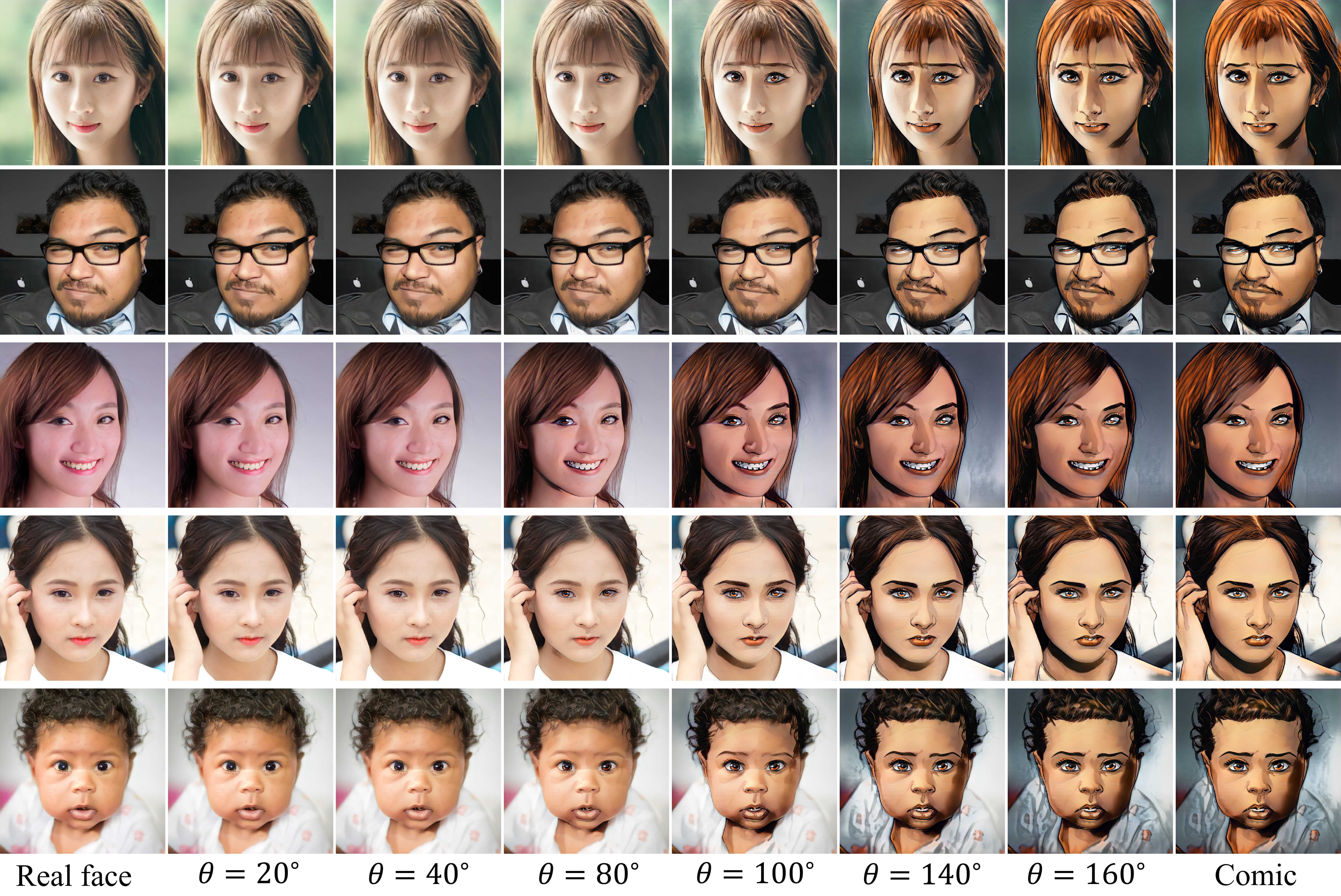}
   \caption{\textbf{Continuous translation from \textit{real face} to \textit{comic}}.}
   \label{fig:face_comic}
\end{figure}

\begin{figure}[t]
  \centering
   \includegraphics[width=1.0\linewidth]{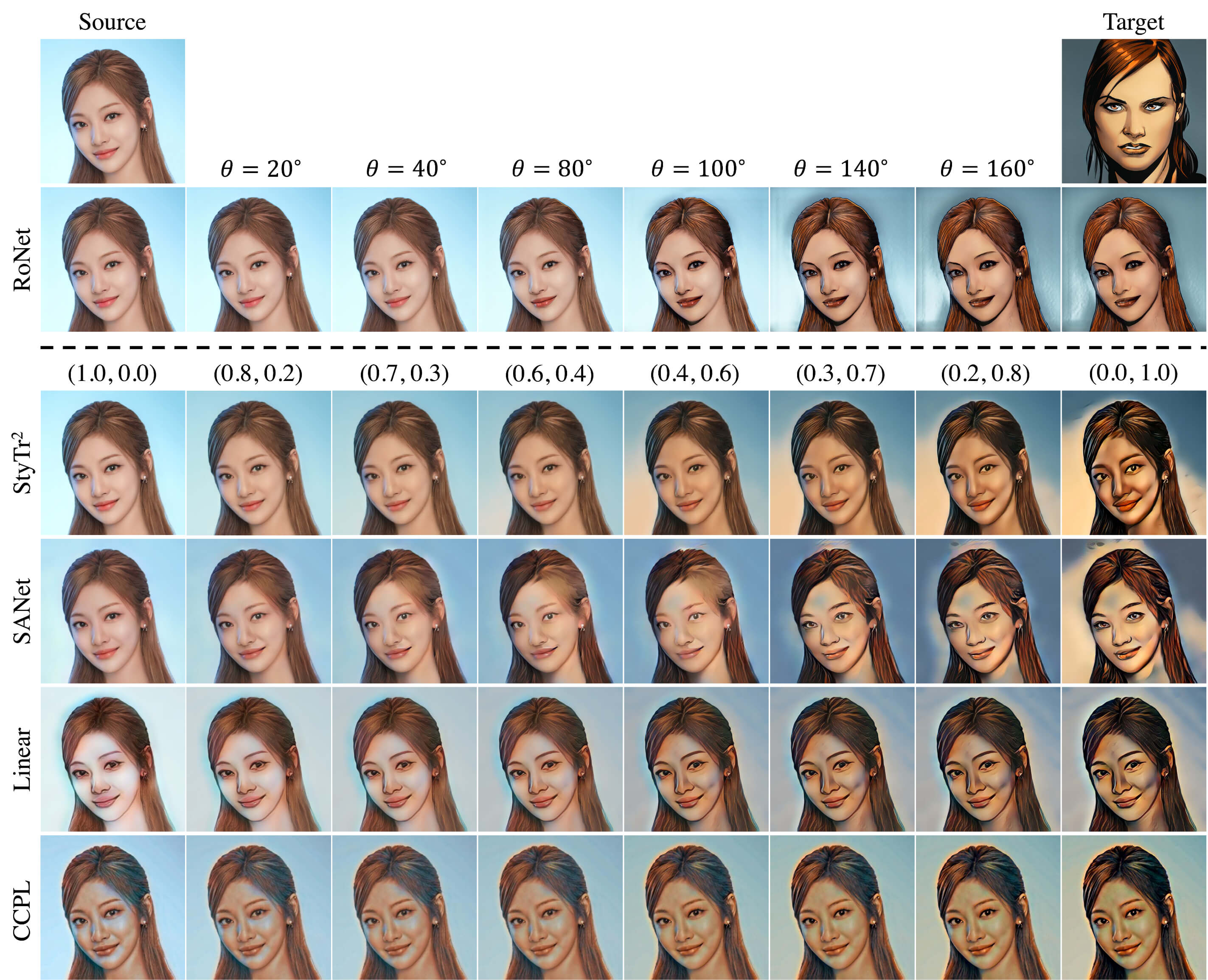}
   \caption{\textbf{Detail comparison of image style transfer results from \textit{real face} to \textit{comic}}.}
   \label{fig:StyleTransfer}
\end{figure}

\begin{figure}[t]
  \centering
   \includegraphics[width=1.0\linewidth]{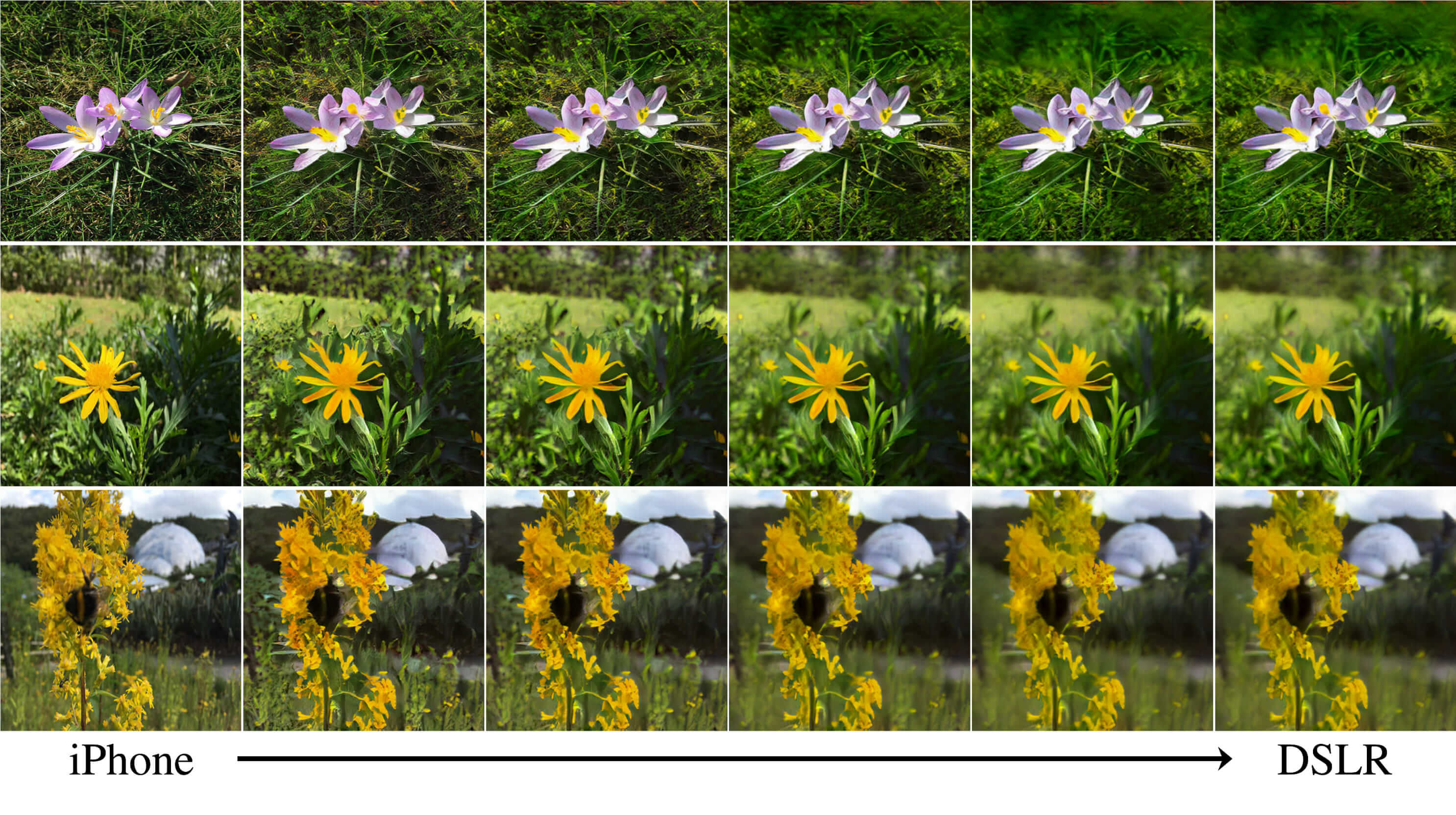}
   \caption{\textbf{Continuous translation on the \textit{iPhone} to \textit{DSLR} task}, using iphone2dslr dataset \cite{zhu2017unpaired}.}
   \label{fig:depth}
\end{figure}

Experiments are also conducted on tasks of $real \; face \to comic \; portrait$ and $iphone \to dslr$, and the visual results are presented in Fig.\ref{fig:face_comic} and Fig.\ref{fig:depth}, respectively.
For \textit{real face} to \textit{comic}, we compared with other four approaches (StyTr \cite{StyTR}, SANet \cite{SANet}, Linear \cite{LST}, and CCPL \cite{wu2022ccpl}) who realize continuous translation with interpolation. From Fig.\ref{fig:StyleTransfer}, it is observed that RoNet yields the most appealing results with rotation and the changes of pupilla color and hairstyle are continuous. The results from \textit{iPhone} to \textit{DSLR} also verify the generalization ability of RoNet.

\begin{figure*}[t]
  \centering
   \includegraphics[width=1.0\linewidth]{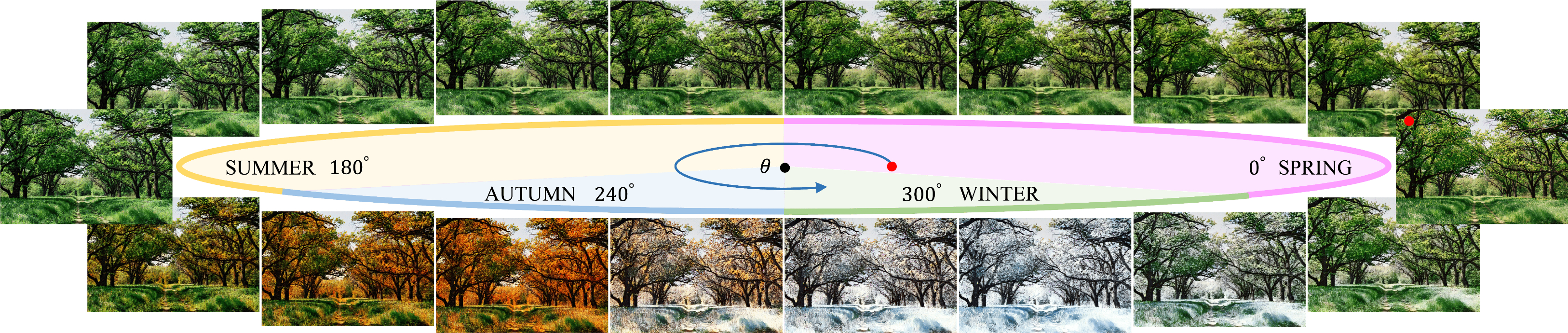}
   \caption{\textbf{The uneven circulation of four seasons} generated by RoNet with the single input (red dot), where the spring and summer are longer than autumn and winter.}
   \label{fig:season_cycle}
\end{figure*}

\begin{figure*}[t]
  \centering
   \includegraphics[width=1.0\linewidth]{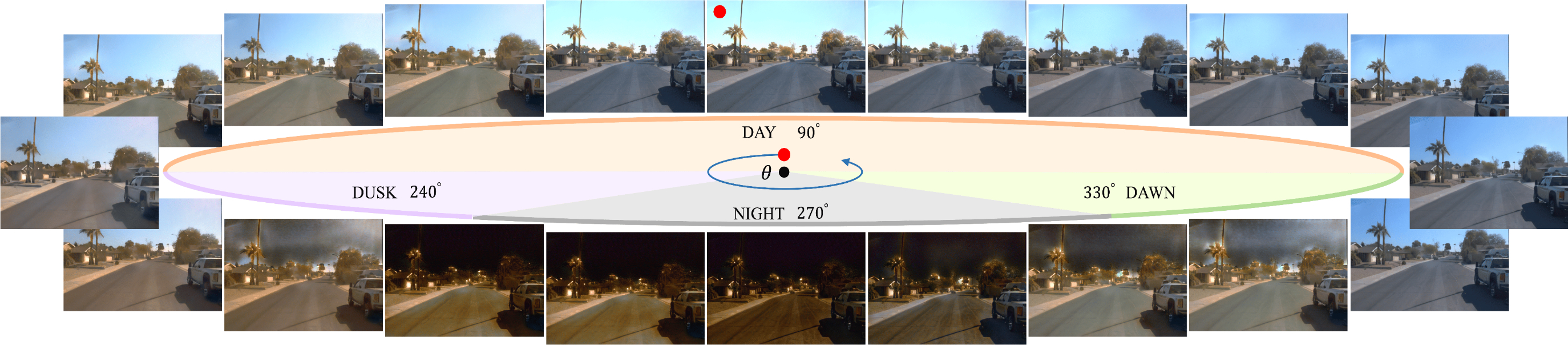}
   \caption{\textbf{Continuous translation of timeshift} where the day and night account for larger part than dusk and dawn.}
   \label{fig:night_cycle}
\end{figure*}

\subsection{Discussion}
\subsubsection{Generalization}
Our proposed method is a general framework that can be applied to most natural scenes with continuous variance, including seasonal circulation and time shifting. However, there are certain special circumstances in the world that need to be considered. For example, in normal situations, the length of day is longer than that of night, dawn and dusk, as shown in Fig.\ref{fig:night_cycle}. Additionally, in the southern hemisphere, the length of spring and summer is longer than that of autumn and winter, as shown in Fig.\ref{fig:season_cycle}. It is important to account for these special circumstances to ensure the effectiveness of our method in such situations.

\subsubsection{Automation} learning imbalanced datasets with label distribution aware margin loss \cite{cao2019learning}

\section{Conclusion}
Aiming at continuous I2I translation, this paper proposes to embed the disentangled style representation under an annular manifold constraint.
And thus the continuous generation can be achieved by rotating the style representation arbitrarily in the proper plane.
To this end, RoNet is designed by implanting a rotation module in the generation network and adding a new patch-based semantic style loss.
Different from the typical linear interpolation, the rotation is capable of moving the style representation from one domain to another with a single input as well as keeping the magnitude of the representation.
Experiments of various translation scenarios (involving seasons, faces, solar days and camera effects) are conducted.
The qualitative and quantitative results demonstrate that our method not only generates the most promising results with plausible transition compared with the others, but also achieves better performance in metrics.
In the future, we plan to study more complex manifolds in continuous translation for better generality.

\bibliographystyle{IEEEtran}
\bibliography{egbib}

\end{document}